\documentclass[letterpaper,12pt]{article}

\usepackage{times}
\usepackage[margin=1in]{geometry}

\usepackage{graphicx}
\DeclareGraphicsExtensions{.pdf,.jpeg,.png}
\usepackage{cite}
\usepackage{amsmath}
\usepackage{amssymb}
\usepackage{amsfonts}
\usepackage{titlesec}
\usepackage{enumerate}
\usepackage[linesnumbered,ruled]{algorithm2e}
\usepackage[justification=centering]{caption}
\titlelabel{\thetitle.\quad}
\titleformat*{\section}{\normalfont\bfseries}
\titleformat*{\subsection}{\normalfont\bfseries}
\titleformat*{\subsubsection}{\normalfont\bfseries}
\titleformat*{\paragraph}{\normalfont\bfseries}
\titleformat*{\subparagraph}{\normalfont\bfseries}

\newtheorem{remark}{Remark}

\usepackage{hyperref}
\hypersetup{hypertex=true,
	colorlinks=true,
	linkcolor=red,
	anchorcolor=blue,
	citecolor=green}
\usepackage[all]{xy}
\usepackage{arydshln}

\begin{document}
\date{}

\title{The Unified Mathematical Framework for IMU Preintegration in Inertial-Aided Navigation System}

\author{Yarong Luo, 
	yarongluo@whu.edu.cn\\
	%Jianlang Hu,
	%hujianlang123@whu.edu.cn\\
	Yang Liu,
	liuyangmail@whu.edu.cn\\
 	Chi Guo,
	guochi@whu.edu.cn\\
%	Shengyong You,
%	shengyongyou@whu.edu.cn\\
%	Jianlang Hu,
%	hujianlang123@whu.edu.cn\\
	Jingnan Liu,
	jnliu@whu.edu.cn\\
 	GNSS Research Center, Wuhan University
}

% The following lines should be within the Author's block
%\footnote{Manuscript received December 15, 2015}

%Authors' information should be left commented for the submittal of the %manuscript.  The comments should be removed for the final submittal
%FirstName LastName \\
%Department of XXXXXXXXXX\\
%University of XXXXXX\\
%Address1\\
%City, State, ZIP, Country \\
%email: email@address.com \\
%\and
%FirstName LastName \\
%Department of XXXXXXXXXX\\
%University of XXXXXX\\
%Address1\\
%City, State, ZIP, Country \\
%email: email@address.com \\
%\and
%FirstName LastName \\
%Department of XXXXXXXXXX\\
%University of XXXXXX\\
%Address1\\
%City, State, ZIP, Country \\
%email: email@address.com \\

\maketitle

\thispagestyle{empty}

\noindent
{\bf\normalsize Abstract}\newline
{This paper proposes a unified mathematical framework for inertial measurement unit (IMU) preintegration in inertial-aided navigation system in different frames under different motion condition. The navigation state is precisely discretized as three parts: local increment, global state, and global increment. The global increment can be calculated in different frames such as local geodetic navigation frame and earth-centered-earth-fixed frame. The local increment which is referred as the IMU preintegration can be calculated under different assumptions according to the motion of the agent and the grade of the IMU.
Thus, it more accurate and more convenient for online state estimation of inertial-integrated navigation system under different environment. Furthermore, the covariance propagation based on left perturbation is proposed for the first time, which is independent of the inputs of the gyroscope and accelerometer. Finally, we show the monotonicity of the uncertainty for determinant optimality criteria and R\'enyi entropy optimality criteria. 
} \vspace{2ex}

\noindent
{\bf\normalsize Key Words}\newline
{pre-integration, $SE_2(3)$ matrix Lie group, inertial-integrated navigation, local increment, equivalent rotation vector, left perturbation, uncertainty propagation, monotonicity preservation}

%%%%%%%%%%%%%%%%%%%%%%%%%%%%%%%%%%%%%%%%%%%%%%%%%%%%%%%%%%%%%%%%%%%%%%%%%%%%%%%%%%%%%%%%%%%%%%%%%%%%%%%%%%%%%%%%%%%%%
%%%%%%%%%%%%%%%%%%%%%%%%%%%%%%%%%%%%%%%%%%%%%%%%%%%%%%%%%%%%%%%%%%%%%%%%%%%%%%%%%%%%%%%%%%%%%%%%%%%%%%%%%%%%%%%%%%%%%
\section{Introduction}
The preintegration theory of inertial measurement unit (IMU) plays a vital role in factor graph based optimization approaches as it integrates IMU measurements at high frequencies in a local frame and is independent of the initial navigation state.
The preintegration theory is introduced by Lupton~\cite{lupton2011visual} and is lifted to the manifold $SO(3)$ by Forster et al.~\cite{forster2016manifold}. However, they assume that the linear acceleration is constant in the world frame which is problematic if the agent is moving fast. The accurate IMU preintegration model is proposed by Henawy et al.~\cite{henawy2021accurate} where the linear acceleration and angular are assumed to be constant between two IMU measurements. Eckenhoff et al.~\cite{eckenhoff2019closed} introduce preintegration theory in continuous form by quaternion which is also based on the piecewise constant IMU measurements assumption. 
Gentil et al.~\cite{le2020gaussian} propose a Gaussian process preintegration for asynchronous inertial-aided state estimation.
Barrau et al.~\cite{brossard2020associating, brossard2021associating} revisit the preintegration theory on matrix Lie group $SE_2(3)$ with rotating Earth.
However, most current methods do not take into account the grade of inertial sensors and the harsh environment in which they are applied. Hence, the preintegration theory on different frames are derived based on the second order kinematic equations of inertial navigation system which are constructed on the matrix Lie group $SE_2(3)$. Furthermore, the multi-sample error compensated algorithms are incorporated into the preintegration theory for high dynamic environment or the vibrating environment.

It is worth noting that the covariance propagation of the error state in all the current methods is dependent on the specific force of the accelerometer. However, the specific force in low-grade inertial sensors may be noisy and prone to large errors~\cite{scherzinger1994modified}. Therefore, the common frame error is used to derive the covariant propagation of the error state and results in a new formula which only depends on the Earth rate and the gravitational constant.  
 
The contributions of the paper can be summarized as follows:\\
1. We derive the preintegration theory on different frames based on matrix Lie group $SE_2(3)$.\\
2. We incorporate the multi-sample error compensated algorithms into the preintegration theory by a local increment calculation. \\
3. We present a new formula for the covariance propagation of the error state that is independent of the inputs of the accelerometer and gyroscope, but depends on the Earth rate and the the gravitational constant.\\
\vspace{-0.1cm}
This remainder of this paper is organized as follows. Preliminaries are presented in Section 2. 
In section 3 the second order kinematic equations of navigation state on matrix Lie group $SE_2(3)$ is derived. In section 4 the exact discrete model of navigation state is introduced and the local increment based on multi-sample error compensated algorithm is derived. Section 5 formulates the preintegration theory on different frames. Section 6 provides the batch and increment recursive formulas for the navigation state and its associated noise. The analytic bias update is derived in Section7. The preintegration measurement residual and Jacobians are presented in Section 8. Section 9 shows that the monotonicity of the uncertainty is preserved for D-opt and R\'enyi entropy. Conclusion and future work are given in Section 10.
%%%%%%%%%%%%%%%%%%%%%%%%%%%%%%%%%%%%%%%%%%%%%%%%%%%%%%%%%%%%%%%%%%%%%%%%%%%%%%%%%%%%%%%%%%%%%%%%%%%%%%%%%%%%%%%%%%%%%
\section{Preliminaries}
The kinematics of the vehicles are described by the velocity, position and the direction, which are expressed on the manifold space and identified by different frames. The velocity and position can be represented by the vectors and the attitude in the 3-dimensional vector space can be represented by the direction cosine matrix (DCM). 
This three quantities can be reformulated as an element of the $SE_2(3)$ matrix Lie group. 
Meanwhile, the vector $v_{ab}^c$ describes the vector points from point a to point b and expressed in the c frame. The direction cosine matrix $C_d^f$ represents the rotation from the d frame to the f frame. Therefore, we summarize the commonly used frames in the inertial navigation and give detailed navigation equations in both the NED frame and the ECEF frame.
\subsection{The $SE_2(3)$ Matrix Lie Group}
The $SE_2(3)$ matrix Lie group is also called the group of direct spatial isometries~\cite{barrau2017the} and it represents the space of matrices that apply a rigid body rotation and 2 translations to points in $\mathbb{R}^3$. Moreover, the group $SE_2(3)$ has the structure of the semidirect product of SO(3) group by $\mathbb{R}^3\times\mathbb{R}^3$ and can be expressed as $SE_2(3)=SO(3)\ltimes \underbrace{\mathbb{R}^3\times\mathbb{R}^3}_2$~\cite{luo2020geometry}. The relationship between the Lie algebra and the associated vector is described by a linear isomorphism $\Lambda$: $\mathbb{R}^9 \rightarrow \mathfrak{se}_2(3)$, i.e. 
\begin{equation}\label{vector_algebra}
\Lambda(\xi)=\begin{bmatrix}
\phi\times & \vartheta &\zeta\\ 0_{1\times3}&0&0\\0_{1\times3}&0&0
\end{bmatrix}\in \mathfrak{se}_2(3), \forall \xi=\begin{bmatrix}
\phi \\ \vartheta \\ \zeta
\end{bmatrix}\in \mathbb{R}^9,\phi,\vartheta,\zeta\in\mathbb{R}^3
\end{equation}

The exponential mapping from the Lie algebra to the corresponding Lie group is given as
\begin{equation}\label{algebra_group}
\begin{aligned}
&T=\exp_G(\Lambda(\xi))=Exp(\xi)=\sum_{n=0}^{\infty}\frac{1}{n!}\left(\Lambda(\xi)\right)^n\\
=&\exp_G\left(\begin{bmatrix}
\phi\times & \vartheta &\zeta\\ 0_{1\times3}&0&0\\0_{1\times3}&0&0
\end{bmatrix} \right)=\begin{bmatrix}
\exp_G(\phi\times) & J\vartheta &J\zeta\\ 0_{1\times3}&1&0\\0_{1\times3}&0&1
\end{bmatrix} 
\end{aligned}
\end{equation}
where $\phi\times$ denotes the skew-symmetric matrix generated from a 3D vector $\phi\in\mathbb{R}^3$; $\exp_G$ denotes the matrix exponential mapping; 
$Exp(\cdot)$ is the composition of $\exp_G$ and $\Lambda$.
$J$ is the left Jacobian matrix of  the 3D orthogonal rotation matrices group $SO(3)$ which is given by:
\begin{equation}\label{left_Jacobian}
J=J_l(\phi)=\sum_{n=0}^{\infty}\frac{1}{(n+1)!}(\phi_{\wedge})^n=I_3+\frac{1-\cos\theta}{\theta^2}\phi_{\wedge}+\frac{\theta-\sin\theta}{\theta^3}\phi_{\wedge}^2,\theta=||\phi||
\end{equation}

The closed form expression for $T$ from the exponential map can also be obtained as
\begin{equation}\label{Exponential_map_se_k_3}
\begin{aligned}
T=\sum_{n=0}^{\infty}\frac{1}{n!}\left(\Lambda(\xi)\right)^n
=I_{5\times 5}+\Lambda(\xi)+\frac{1-\cos\theta}{\theta^2}\Lambda(\xi)^2+\frac{\theta-\sin\theta}{\theta^3}\Lambda(\xi)^3
\end{aligned}
\end{equation}

$SE_2(3)$ is commonly used as the extended poses (orientation, velocity, position) for 3-dimensional inertial navigation. 

Then, a useful auxiliary function introduce by ~\cite{bloesch2013state} is given
\begin{equation}\label{Gamma}
\Gamma_m(\phi)\triangleq \sum_{n=0}^{\infty}\frac{1}{(n+m)!}(\phi_{\wedge}^n)
\end{equation}
Then the integrals can be easily expressed and computed by the matrix Taylor series 
\begin{equation}\label{Gamma_0}
\Gamma_0(\phi)=I_3+\frac{\sin||\phi||}{||\phi||}\phi_{\wedge}+\frac{1-\cos||\phi||}{||\phi||^2}\phi_{\wedge}^2=T
\end{equation}
\begin{equation}\label{Gamma_11}
\Gamma_1=I_3+\frac{1-\cos||\phi||}{||\phi||^2}\phi_{\wedge}+\frac{||\phi||-\sin||\phi||}{||\phi||^3}\phi_{\wedge}^2=J=J_l(\phi)
\end{equation}
It is worth noting that $\Gamma_0(\phi)$ is the exponential mapping of $SO(3)$, while $\Gamma_1(\phi)$ is the left Jacobian of $SO(3)$~\cite{hartley2020contact}.

Since we have  $\Gamma_2(\phi)\phi_{\wedge}+I_3=\Gamma_1(\phi)$ and $\Gamma_2(\phi)$ can be represented as $\Gamma_2(\phi)=\frac{1}{2!}I_3+x\phi_{\wedge}+y\phi_{\wedge}^2$, then $x$ and $y$ can be obtained by determined coefficient method:
\begin{equation}\label{Gamma_22}
\Gamma_2(\phi)=\frac{1}{2}I_3+\frac{||\phi||-\sin||\phi||}{||\phi||^3}\phi_{\wedge}+\frac{||\phi||^2+2\cos||\phi||-2}{2||\phi||^4}\phi_{\wedge}^2
\end{equation}

Similarly, as we have $\Gamma_3(\phi)\phi_{\wedge}+\frac{1}{2}I_3=\Gamma_2(\phi)$ and $\Gamma_3(\phi)$ can be represented as $\Gamma_3(\phi)=\frac{1}{3!}I_3+a\phi_{\wedge}+b\phi_{\wedge}^2$, then $a$ and $b$ can be obtained by determined coefficient method:
\begin{equation}\label{Gamma_3}
\Gamma_3(\phi)=\frac{1}{3!}I_3+\frac{||\phi||^2+2\cos||\phi||-2}{2||\phi||^4}\phi_{\wedge}+\frac{||\phi||^3-6||\phi||+6\sin||\phi||}{6||\phi||^5}\phi_{\wedge}^2
\end{equation}

It can be verified that 
\begin{equation}\label{left_Jacobian_to_right_Jacobian}
\Gamma_{m}(-\phi)=\Gamma_{m}(\phi)^T
\end{equation}
and
\begin{equation}\label{rotation_for_all_left_and_right}
\Gamma_{m}\left(\Gamma_{0}(\phi)\phi\right)=\Gamma_{0}(\phi)\Gamma_{m}(\phi)\Gamma_{0}(-\phi)\Rightarrow \Gamma_{0}(\phi)\Gamma_{m}(\phi)=\Gamma_{m}(\Gamma_{0}(\phi)\phi)\Gamma_{0}(\phi)
\end{equation}

The linearization of a function $\Gamma_{m}(\cdot )$ is the first order Taylor series of the function evaluated as a certain element of the domain. 
If we assume that $\phi$ is small, then using the first order Taylor series we can obtain
\begin{equation}\label{BCH_left}
\Gamma_{m}(\phi+\psi)\approx Exp(\Gamma_{m+1}(\psi)\phi)\Gamma_{m}(\psi)=\Gamma_{0}(\Gamma_{m+1}(\psi)\phi)\Gamma_{m}(\psi)
\end{equation}
It is obvious that the approximate BCH formula that using left Jacobian matrix is the case when $m=0$.

If we assume that $\psi$ is small, then using the first order Taylor series we can obtain
\begin{equation}\label{BCH_1}
\Gamma_{m}(\phi+\psi)\approx \Gamma_{m}(\phi)Exp(\Gamma_{m+1}(-\phi)\psi)=\Gamma_{m}(\phi)\Gamma_{0}(\Gamma_{m+1}(-\phi)\psi)
\end{equation}
It is obvious that the approximate BCH formula that using right Jacobian matrix is the case when $m=0$.
\subsection{Uncertainty and Concentrated Gaussian Distribution on Matrix Lie Group $SE_2(3)$} 
The uncertainties on matrix Lie group $SE_2(3)$ can be represented by left multiplication and right multiplication
\begin{equation}\label{uncertainty}
\begin{aligned}
\text{left multiplication}:{T}_l&=\hat{T}\exp_G(\Lambda{(\varepsilon_l)})=\hat{T}\exp_G(\Lambda{(\varepsilon_l)})\hat{T}^{-1}\hat{T}=\exp_G(\Lambda{(Ad_{\hat{T}}(\varepsilon_l))})\hat{T}\\
\text{right multiplication}:{T}_r&=\exp_G(\Lambda{(\varepsilon_r)})\hat{T}
\end{aligned}
\end{equation}

Therefore, the probability distributions for the random variables $T\in SE_2(3)$ can be defined as left-invariant concentrated Gaussian distribution on $SE_2(3)$ and right-invariant concentrated Gaussian distribution on $SE_2(3)$:
\begin{equation}\label{concentrated}
\begin{aligned}
\text{left-invariant}:T\sim \mathcal{N}_L(\hat{T},P),{T}_l&=\hat{T}\exp_G(\Lambda{(\varepsilon_l)}),\varepsilon_l\sim \mathcal{N}(0,P)\\
\text{right-invariant}:T\sim  \mathcal{N}_R(\hat{T},P),{T}_r&=\exp_G(\Lambda{(\varepsilon_r)})\hat{T},\varepsilon_r\sim   \mathcal{N}(0,P)
\end{aligned}
\end{equation}
where $\mathcal{N}(\cdot,\cdot)$ is the classical Gaussian distribution in Euclidean space and $P\in\mathbb{R}^{3(K+1)\times 3(K+1)}$ is a covariance matrix. The invariant property can be verified by $\exp_G(\Lambda{(\varepsilon_r)})=({T}_r \Gamma) (\hat{T}\Gamma)^{-1}={T}_r \hat{T}^{-1}$ and $\exp_G(\Lambda{(\varepsilon_l)})=(\Gamma\hat{T})^{-1}(\Gamma{T}_l ) =\hat{T}^{-1}{T}_l$. The noise-free quantity $\hat{T}$ is viewed as the mean, and the dispersion arises through left multiplication or right multiplication with the matrix exponential of a zero mean Gaussian random variable.
\subsection{Reference Frames}
The commonly used reference frames~\cite{shin2005estimation} in inertial-integrated navigation system are summarized in following. 

Earth-Centered-Inertial (ECI) Frames (i-frame) is an ideal frame of reference in which ideal accelerometers and gyroscopes fixed to the i-frame have zero outputs and it has its origin at the center of the Earth and axes that are non-rotating with respect to the fixed stars with its z-axis parallel to the spin axis of the Earth, x-axis pointing towards the mean vernal equinox, and y-axis completing a right-handed orthogonal frame.

Earth-Centered-Earth-Fixed (ECEF) Frames (e-frame) has its origin at the center of mass of the Earth and axes that are fixed with respect to the Earth. Its x-axis points towards the mean meridian of Greenwich, z-axis is parallel to the mean spin axis of the Earth, and y-axis completes a right-handed orthogonal frame.

Navigation Frames (n-frame) is a local geodetic frame which has its origin coinciding with that of the sensor frame, with its x-axis pointing towards geodetic north, z-axis orthogonal to the reference ellipsoid pointing down, and y-axis completing a right-handed
orthogonal frame, i.e. the north-east-down (NED) system. The local geodetic coordinate system can be represented by north coordinate X, east coordinate Y and height Z (XYZ, units:m, m, m), or by latitude $\varphi$, longitude $\lambda$ and height $h$ (LLH, unit: rad, rad, m), and longitude and latitude can be converted one-to-one to XY.

Body Frames (b-frame) is an orthogonal axis set which is fixed onto the vehicle and rotate with it, therefore, it is aligned with the roll, pitch and heading axes of a vehicle, i.e. forward-transversal-down.
\subsection{NED Navigation Equations when position is represented in terms of LLH}
The attitude in the NED frame can be represented by the DCM $C_b^n$, 
The differential equation of $C_b^n$ and $C_n^b$ are given by
\begin{equation}\label{C_b_n_d_e}
\dot{C}_b^n=C_b^n(\omega_{ib}^b\times)-(\omega_{in}^n\times)C_b^n
\end{equation}
\begin{equation}\label{C_n_b_d_e}
\dot{C}_n^b=C_n^b(\omega_{in}^n\times)-(\omega_{ib}^b\times)C_n^b
\end{equation}
where $\omega_{ib}^b$ is the angular rate vector of the body frame relative to the inertial frame expressed in the body frame; $\omega_{in}^n$ is the angular rate vector of the navigation frame relative to the inertial frame expressed in the navigation frame.

The differential equation of the velocity vector in the NED local-level navigation frame is given by
\begin{equation}\label{v_eb_n_d_e}
\dot{v}_{eb}^n=C_b^nf_{ib}^b-\left[ (2\omega_{ie}^n+\omega_{en}^n)\times\right]v_{eb}^n+g_{ib}^n
\end{equation}
where $\omega_{ie}^n$ is the earth rotation vector expressed in the navigation frame; $f_{ib}^b$ is the specific force vector in navigation frame; $\omega_{en}^n=\omega_{in}^n-\omega_{ie}^n$ is the angular rate vector of the navigation frame relative to the earth frame expressed in the navigation frame which is also call the transport rate; and $g_{ib}^n$ is the gravity vector.
$\omega_{ie}^n$ and $\omega_{en}^n$ can be given as follows
\begin{equation}\label{omega_ie_n}
\begin{aligned}
&\omega_{ie}^n=\begin{bmatrix}
\omega_{ie}\cos\varphi\\0\\-\omega_{ie}\sin\varphi
\end{bmatrix},\omega_{en}^n=\begin{bmatrix}
\dot{\lambda}\cos\varphi \\-\dot{\varphi} \\ -\dot{\lambda}\sin\varphi
\end{bmatrix}=\begin{bmatrix}
\frac{v_E}{R_N+h}\\ \frac{-v_N}{R_M+h}\\ \frac{-v_E\tan\varphi}{R_N+h}
\end{bmatrix}\\
&\omega_{in}^n=\omega_{ie}^n+\omega_{en}^n=\begin{bmatrix}
\omega_{ie}\cos\varphi+\frac{v_E}{R_N+h}\\ \frac{-v_N}{R_M+h}\\ -\omega_{ie}\sin\varphi -\frac{v_E\tan\varphi}{R_N+h}
\end{bmatrix},2\omega_{ie}^n+\omega_{en}^n=\begin{bmatrix}
2\omega_{ie}\cos\varphi+\frac{v_E}{R_N+h}\\ \frac{-v_N}{R_M+h}\\ -2\omega_{ie}\sin\varphi -\frac{v_E\tan\varphi}{R_N+h}
\end{bmatrix}
\end{aligned}
\end{equation}
where $\omega_{ie}=0.000072921151467rad/s$ is the magnitude of the earth's rotation angular rate; $v_N$ and $v_E$ are velocities in the north and east direction, respectively; $h$ is ellipsoidal height; $R_M$ and $R_N$ are radii of curvature in the meridian and prime vertical; $\dot{\varphi}=\frac{v_N}{R_M+h}$ and $\dot{\lambda}=\frac{v_E}{(R_N+h)\cos\varphi}$ are used in the derivation.

\subsection{NED Navigation Equations  when position is represented in terms of XYZ}
The position vector differential equation in terms of the NED coordinate system can be calculated as
\begin{equation}\label{r_eb_n_d_e}
\dot{r}_{eb}^n=\frac{d}{dt}(C_e^n r_{eb}^e)=\frac{d}{dt}(C_e^n )r_{eb}^e+C_e^n\dot{r}_{eb}^e=C_e^n(\omega_{ne}^e\times)r_{eb}^e+C_e^nv_{eb}^e=-\omega_{en}^n\times r_{eb}^n+v_{eb}^n
\end{equation}

%Perturbations on $\omega_{ie}^n$, $\omega_{en}^n$, and $\omega_{in}^n$ can be given as follows
%\begin{equation}\label{perturbation_omega_ie_n}
%\delta\omega_{ie}^n=\begin{bmatrix}
%\frac{-\omega_{ie}\sin\varphi \delta r_N}{R_M+h} \\0 \\ \frac{-\omega_{ie}\cos\varphi\delta r_N}{R_M+h}
%\end{bmatrix}=M_1\delta r_{eb}^n
%\end{equation}
%\begin{equation}\label{perturbation_omega_en_n}
%\delta\omega_{en}^n=\begin{bmatrix}
%\frac{v_E\delta r_D}{(R_N+h)^2}+\frac{\delta v_E}{R_N+h}\\
%-\frac{v_N\delta r_D}{(R_M+h)^2}-\frac{\delta v_N}{R_M+h}\\
%-\frac{v_E\delta r_N}{(R_N+h)(R_M+h)\cos^2\varphi}-\frac{v_E\tan\varphi\delta r_D}{(R_N+h)^2}-\frac{\tan\varphi \delta v_E}{R_N+h}
%\end{bmatrix}=M_3\delta r_{eb}^n+M_2\delta v_{eb}^n
%\end{equation}
%\begin{equation}\label{perturbation_omega_in_n}
%\delta\omega_{in}^n=\delta\omega_{ie}^n+\delta\omega_{en}^n=M_1\delta r_{eb}^n+M_3\delta r_{eb}^n+M_2\delta v_{eb}^n=(M_1+M_3)\delta r_{eb}^n+M_2\delta v_{eb}^n
%\end{equation}
\subsection{The gravitational vectors in different frames}
The gravitational vector in ECI frame is given as
\begin{equation}\label{gravitational_vector_i}
g_{ib}^i=G_{ib}^i-(\omega_{ie}^i\times)^2r_{eb}^i
\end{equation}
where $g_{ib}^i$ is the gravity vector expressed in ECI frame; $G_{ib}^i$ is the gravitational vector expressed in the ECI frame.

According to equation(\ref{gravitational_vector_i}) we can get The gravitational vector in ECEF frame is given as
\begin{equation}\label{gravitational_vector_e}
g_{ib}^e=C_i^eg_{ib}^i=C_i^eG_{ib}^i-C_i^e(\omega_{ie}^i\times)C_e^iC_i^e(\omega_{ie}^i\times)C_e^iC_i^er_{eb}^i=G_{ib}^e-(\omega_{ie}^e\times)^2r_{eb}^e
\end{equation}
where $g_{ib}^e$ is the gravity vector expressed in ECEF frame; $G_{ib}^e$ is the gravitational vector expressed in the ECEF frame.

According to equation(\ref{gravitational_vector_i}) we can get The gravitational vector in navigation frame is given as
\begin{equation}\label{gravitational_vector_n}
g_{ib}^n=C_i^ng_{ib}^i=C_i^nG_{ib}^i-C_i^n(\omega_{ie}^i\times)C_n^iC_i^n(\omega_{ie}^i\times)C_n^iC_i^nr_{eb}^i=G_{ib}^n-(\omega_{ie}^n\times)^2r_{eb}^n
\end{equation}
where $g_{ib}^n$ is the gravity vector expressed in navigation frame; $G_{ib}^n$ is the gravitational vector expressed in the navigation frame.
\subsection{ECEF Navigation Equations with position is represented as XYZ}
The differential equation of the attitude matrix in the ECEF frame can be represented as
\begin{equation}\label{C_b_e_d_e}
\dot{C}_b^e=C_b^e(\omega_{ib}^b\times)-(\omega_{ie}^e\times)C_b^e
\end{equation}
\begin{equation}\label{C_e_b_d_e}
\dot{C}_e^b=C_e^b(\omega_{ie}^e\times)-(\omega_{ib}^b\times)C_e^b
\end{equation}

The differential equation of the velocity vector in the ECEF frame is given as
\begin{equation}\label{v_eb_e_d_e}
\dot{v}_{eb}^e=C_b^ef_{ib}^b-2\omega_{ie}^e\times v_{eb}^e+g_{ib}^e
\end{equation}

The differential equation of the position vector in the ECEF frame is given as
\begin{equation}\label{r_eb_e_d_e}
\dot{r}_{eb}^e=v_{eb}^e
\end{equation}
\subsection{Another ECEF Navigation Equations with position is represented as XYZ}
	As the ECEF frame has the same origin as the ECI frame, so $r_{ie}^i=0$ and $r_{ib}^i=r_{ie}^i+r_{eb}^i=r_{eb}^i=C_e^ir_{eb}^e$. Meanwhile, we also get $r_{ib}^e=r_{eb}^e=C_i^er_{eb}^i$.

The differential equation of the attitude $C_b^e$ is given as
\begin{equation}\label{ECEF_ground_nonattitude}
\dot{C}_b^e=C_b^e(\omega_{ib}^b\times)-(\omega_{ie}^e\times)C_b^e
\end{equation}

As the velocity has the relationship $v_{ib}^e=C_i^ev_{ib}^i$, so the differential equation of the velocity $v_{ib}^e$ can be calculated as
\begin{equation}\label{ECEF_ground_nonvelocity}
\begin{aligned}
\dot{v}_{ib}^e&=\frac{d}{dt}(C_i^ev_{ib}^i)=\dot{C}_i^ev_{ib}^i+C_i^e\dot{v}_{ib}^i=(-\omega_{ie}^e\times){C}_i^ev_{ib}^i+C_i^e\left(C_b^if_{ib}^b+G_{ib}^i\right)\\
&=(-\omega_{ie}^e\times)v_{ib}^e+C_i^eC_b^if_{ib}^b+C_i^eG_{ib}^i=(-\omega_{ie}^e\times)v_{ib}^e+C_b^ef_{ib}^b+G_{ib}^e
\end{aligned}
\end{equation}
where $G_{ib}^e$ is the gravity acceleration expressed in the ECEF frame.

The differential equation of the position ${r}_{ib}^e$ is given as
\begin{equation}\label{ECEF_ground_nonpostion}
v_{eb}^e=\dot{r}_{eb}^e=\dot{r}_{ib}^e=(-\omega_{ie}^e\times)C_i^er_{ib}^i+C_i^e\dot{r}_{ib}^i=(-\omega_{ie}^e\times)r_{ib}^e+v_{ib}^e
\end{equation}

According to the differential equation of position (\ref{ECEF_ground_nonpostion}) we can know that $v_{ib}^e=v_{eb}^e+(\omega_{ie}^e\times)r_{ib}^e$, so the differential equation of velocity $v_{ib}^e$ can also be deduced as follows:
\begin{equation}\label{ECEF_non_ground_gravity_velocity1}
\dot{v}_{ib}^e=\dot{v}_{eb}^e+(\omega_{ie}^e\times)\dot{r}_{ib}^e
\end{equation}

Substituting equation(\ref{gravitational_vector_e}) into equation(\ref{ECEF_non_ground_gravity_velocity1}) and we can get
\begin{equation}\label{ECEF_non_ground_gravity_velocity2}
\begin{aligned}
\dot{v}_{ib}^e&=C_b^ef_{ib}^b+g_{ib}^e-2(\omega_{ie}^e\times)v_{eb}^e+(\omega_{ie}^e\times)\dot{r}_{eb}^e=C_b^ef_{ib}^b+g_{ib}^e-(\omega_{ie}^e\times)v_{eb}^e\\
&=C_b^ef_{ib}^b+g_{ib}^e-(\omega_{ie}^e\times)((-\omega_{ie}^e\times)r_{ib}^e+v_{ib}^e)\\
&=C_b^ef_{ib}^b+g_{ib}^e+(\omega_{ie}^e\times)^2r_{ib}^e-(\omega_{ie}^e)\times v_{ib}^e=C_b^ef_{ib}^b+G_{ib}^e-(\omega_{ie}^e)\times v_{ib}^e
\end{aligned}
\end{equation}
This result is the same as the equation(\ref{ECEF_ground_nonvelocity}).

In the end, we get different differential equations of the attitude, velocity and the position in the ECEF frame.

\subsection{Sensor Error Modeling}
The sensor errors of the accelerometers and gyroscopes for consumer-grade inertial measurement unit (IMU) are modeled as one-order Gauss-Markov model:
\begin{equation}\label{accelerometers_bias}
\delta f_{ib}^b=b_a+w_a, \dot{b}_a=-\frac{1}{\tau_a}b_a+w_{b_a}
\end{equation}
\begin{equation}\label{gyroscopes_bias}
\delta \omega_{ib}^b=b_g+w_g, \dot{b}_g=-\frac{1}{\tau_g}b_g+w_{b_g}
\end{equation}
where $w_a$ and $w_g$ are the Gaussian white noises of the accelerometers and gyroscopes, respectively; $w_{b_a}$  and  $w_{b_g}$  are the Gaussian white noises of the accelerometer biases and gyroscope biases, respectively; $\tau_a$ and $\tau_g$ are the correlation times of accelerometer biases and gyroscope biases, respectively.

Of course, the sensor errors of accelerometers and gyroscopes can also be modeled as random walk process for intermediate-grade IMU and the navigation-grade IMU:
\begin{equation}\label{accelerometers_bias_rw}
\delta f_{ib}^b=b_a+w_a, \dot{b}_a=0
\end{equation}
\begin{equation}\label{gyroscopes_bias_rw}
\delta \omega_{ib}^b=b_g+w_g, \dot{b}_g=0
\end{equation}
%%%%%%%%%%%%%%%%%%%%%%%%%%%%%%%%%%%%%%%%%%%%%%%%%%%%%%%%%%%%%%%%%%%%%%%%%%%%%%%%%%%%
\section{Kinematic Equations for Inertial Navigation System}\label{sectionIII}
\subsection{Kinematic Equations on NED frame}
As the position can be represented in terms of LLH and XYZ, different right $SE_2(3)$ based EKF for NED navigation is derived in this section. And more derivations can be found in~\cite{luo2021se23} as the core idea is the same.
The velocity vector $v_{eb}^n$, position vector $r_{eb}^n$ and attitude matrix $C_b^n$ can formula the element of the $SE_2(3)$ matrix Lie group

\begin{equation}\label{matrix_Lie_group_eb_n}
\mathcal{X}=\begin{bmatrix}
C_b^n & v_{eb}^n & r_{eb}^n\\
0_{1\times3} & 1 & 0\\
0_{1\times 3} & 0& 1
\end{bmatrix}\in SE_2(3)
\end{equation}

The inverse of the element can be written as follows
\begin{equation}\label{inverse_matrix_Lie_group_eb_b}
\mathcal{X}^{-1}=\begin{bmatrix}
C_n^b & -C_n^bv_{eb}^n & -C_n^br_{eb}^n\\
0_{1\times 3} &1 &0\\
0_{1\times 3} &0&1
\end{bmatrix}=\begin{bmatrix}
C_n^b & -v_{eb}^b & -r_{eb}^b\\
0_{1\times 3} &1 &0\\
0_{1\times 3} &0&1
\end{bmatrix}\in SE_2(3)
\end{equation}

The INS mechanization in NED frame in terms of XYZ is given as
\begin{equation}\label{C_b_n_d_e_invariant}
	\dot{C}_b^n=C_b^n(\omega_{ib}^b\times)-(\omega_{in}^n\times)C_b^n
\end{equation}
\begin{equation}\label{v_eb_n_d_e_invarant}
	\dot{v}_{eb}^n=C_b^nf_{ib}^b-\left[ (2\omega_{ie}^n+\omega_{en}^n)\times\right]v_{eb}^n+g_{ib}^n
\end{equation}
\begin{equation}\label{r_eb_n_d_e_invariant}
	\dot{r}_{eb}^n=\frac{d}{dt}(C_e^n r_{eb}^e)=\frac{d}{dt}(C_e^n )r_{eb}^e+C_e^n\dot{r}_{eb}^e=C_e^n(\omega_{ne}^e\times)r_{eb}^e+C_e^nv_{eb}^e=-\omega_{en}^n\times r_{eb}^n+v_{eb}^n
\end{equation}
where $g_{ib}^n$ is the gravity vector, and its relationship with the gravitational vector $\overline{g}_{ib}^n$ is given by
\begin{equation}\label{gravitation_gravity}
	g_{ib}^n=\overline{g}_{ib}^n-(\omega_{ie}^n\times)^2r_{eb}^n
\end{equation}

Therefore, the differential equation of the $\mathcal{X}$ can be calculated as
\begin{equation}\label{differential}
\begin{aligned}
&\frac{d}{dt}\mathcal{X}=f_{u_t}(\mathcal{X})=\frac{d}{dt}\begin{bmatrix}
C_b^n & v_{eb}^n & r_{eb}^n\\
0_{1\times3} & 1 & 0\\
0_{1\times 3} & 0& 1
\end{bmatrix}
=\begin{bmatrix}
\dot{C}_b^n & \dot{v}_{eb}^n & \dot{r}_{eb}^n\\
0_{1\times3} & 0 & 0\\
0_{1\times 3} & 0& 0
\end{bmatrix}=\mathcal{X}W_1+W_2\mathcal{X}\\
=&\begin{bmatrix}
C_b^n(\omega_{ib}^b\times)-(\omega_{in}^n\times)C_b^n & C_b^nf_{ib}^b-\left[ (2\omega_{ie}^n+\omega_{en}^n)\times\right]v_{eb}^n+g_{ib}^n & -\omega_{en}^n\times r_{eb}^n+v_{eb}^n\\
0_{1\times3} & 0 & 0\\
0_{1\times 3} & 0& 0
\end{bmatrix}
\end{aligned}
\end{equation}
where $u_t$ is a sequence of inputs; $W_1$ and $W_2$ are denoted as
\begin{equation}\label{W_1_W_2}
W_1=\begin{bmatrix}
\omega_{ib}^b\times & f_{ib}^b & 0\\
0_{1\times3} & 0 & 0\\
0_{1\times 3} & 0& 0
\end{bmatrix},W_2=\begin{bmatrix}
-\omega_{in}^n\times & g_{ib}^n-\omega_{ie}^n\times v_{eb}^n & v_{eb}^n+\omega_{ie}^n\times r_{eb}^n\\
0_{1\times3} & 0 & 0\\
0_{1\times 3} & 0& 0
\end{bmatrix}
\end{equation}

It is easy to verify that the dynamical equation $f_{u_t}(\mathcal{X})$ is group-affine and the group-affine system owns the log-linear property of the corresponding error propagation~\cite{barrau2017the}:
\begin{equation}\label{proof_invariance}
\begin{aligned}
&f_{u_t}(\mathcal{X}_A)\mathcal{X}_B+\mathcal{X}_Af_{u_t}(\mathcal{X}_B)-\mathcal{X}_Af_{u_t}(I_d)\mathcal{X}_B\\
=&(\mathcal{X}_AW_1+W_2\mathcal{X}_A)\mathcal{X}_B+\mathcal{X}_A(\mathcal{X}_BW_1+W_2\mathcal{X}_B)-\mathcal{X}_A(W_1+W_2)\mathcal{X}_B\\
=&\mathcal{X}_A\mathcal{X}_BW_1+W_2\mathcal{X}_A\mathcal{X}_B\triangleq f_{u_t}(\mathcal{X}_A\mathcal{X}_B)
\end{aligned}
\end{equation}
\subsection{Kinematic Equations on transformed NED Frame}
\label{transformed_INS}
Similar to\cite{brossard2020associating}, an auxiliary velocity is introduced as
\begin{equation}\label{auxiliary_velocity}
	\overline{v}_{eb}^n=v_{eb}^n+\omega_{ie}^n\times r_{eb}^n=v_{eb}^n+C_e^n\omega_{ie}^e\times r_{eb}^n
\end{equation}

With the introduced auxiliary velocity vector, the transformed INS mechanization is now given by
\begin{equation}\label{C_b_n_d_e_invariant1}
	\dot{C}_b^n=C_b^n(\omega_{ib}^b\times)-(\omega_{in}^n\times)C_b^n
\end{equation}
\begin{equation}\label{v_eb_n_d_e_invarant1}
	\dot{v}_{eb}^n=C_b^nf_{ib}^b-(\omega_{in}^n)\times \overline{v}_{eb}^n+\overline{g}^n
\end{equation}
\begin{equation}\label{r_eb_n_d_e_invariant1}
	\dot{r}_{eb}^n=-\omega_{in}^n\times r_{eb}^n+\overline{v}_{eb}^n
\end{equation}

Then defining the state composed by the attitude $C_b^n$, the velocity $\overline{v}_{eb}^n$, and the position $r_{eb}^n$ as the elements of the matrix Lie group $SE_2(3)$, that is
\begin{equation}\label{new_state}
	\mathcal{X}=\begin{bmatrix}
		C_b^n & \overline{v}_{eb}^n & r_{eb}^n\\
		0_{1\times 3} & 1 &0\\
		0_{1\times 3} & 0 & 1
	\end{bmatrix}
\end{equation}

Therefore, equation(\ref{C_b_n_d_e_invariant1}), equation(\ref{v_eb_n_d_e_invarant1}), equation(\ref{r_eb_n_d_e_invariant1}) can be rewritten in a compact form as
\begin{equation}\label{differential_invariant}
	\begin{aligned}
		&\frac{d}{dt}\mathcal{X}=f_{u_t}(\mathcal{X})=\frac{d}{dt}\begin{bmatrix}
			C_b^n & \overline{v}_{eb}^n & r_{eb}^n\\
			0_{1\times3} & 1 & 0\\
			0_{1\times 3} & 0& 1
		\end{bmatrix}
		=\begin{bmatrix}
			\dot{C}_b^n & \dot{\overline{v}}_{eb}^n & \dot{r}_{eb}^n\\
			0_{1\times3} & 0 & 0\\
			0_{1\times 3} & 0& 0
		\end{bmatrix}=\mathcal{X}W_1+W_2\mathcal{X}\\
		=&\begin{bmatrix}
			C_b^n(\omega_{ib}^b\times)-(\omega_{in}^n\times)C_b^n & C_b^nf_{ib}^b-(\omega_{in}^n)\times \overline{v}_{eb}^n+\overline{g}^n & -\omega_{in}^n\times r_{eb}^n+\overline{v}_{eb}^n\\
			0_{1\times3} & 0 & 0\\
			0_{1\times 3} & 0& 0
		\end{bmatrix}
	\end{aligned}
\end{equation}
where $W_1$ and $W_2$ are denoted as
\begin{equation}\label{W_1_W_2_invariant}
	W_1=\begin{bmatrix}
		\omega_{ib}^b\times & f_{ib}^b & 0\\
		0_{1\times3} & 0 & 0\\
		0_{1\times 3} & 0& 0
	\end{bmatrix},W_2=\begin{bmatrix}
		-\omega_{in}^n\times & \overline{g}^n & \overline{v}_{eb}^n\\
		0_{1\times3} & 0 & 0\\
		0_{1\times 3} & 0& 0
	\end{bmatrix}
\end{equation}
It is easy to verify that the dynamical equation(\ref{differential_invariant}) satisfies the group-affine property so that the error state dynamical equation is independent of the global state.
%%%%%%%%%%%%%%%%%%%%%%%%%%%%%%%%%%%%%%%
\subsection{Kinematic Equations on ECEF Navigation}
When the system  state is defined as 
	\begin{equation}\label{new_state_eb_e}
\mathcal{X}=\begin{bmatrix}
C_b^e & v_{eb}^e & r_{eb}^e\\
0_{1\times 3} & 1 & 0\\
0_{1\times 3} & 0& 1
\end{bmatrix}\in SE_2(3)
\end{equation}
	where $C_b^e$ is the direction cosine matrix from the body frame to the ECEF frame; $v_{eb}^e$ is the velocity of body frame relative to the ECEF frame expressed in the ECEF frame; $r_{eb}^e$ is the position of body frame relative to the ECEF frame expressed in the ECEF frame.
	
	Then the dynamic equation of the state $\mathcal{X}$ can be deduced as follows
		\begin{equation}\label{lie_group_state_differential_eb_e}
	\begin{aligned}
	&\frac{d}{dt}\mathcal{X}=f_{u_t}(\mathcal{X})=\frac{d}{dt}\begin{bmatrix}
	C_b^e & v_{eb}^e & r_{eb}^e\\0_{1\times 3}&1&0\\0_{1\times 3}&0&1
	\end{bmatrix}=\begin{bmatrix}
	\dot{C}_b^e & \dot{v}_{eb}^e & \dot{r}_{eb}^e\\0_{1\times 3}&0&0\\0_{1\times 3}&0&0
	\end{bmatrix}\\
	=&\begin{bmatrix}
	C_b^e(\omega_{ib}^b\times)-(\omega_{ie}^e\times)C_b^e & (-2\omega_{ie}^e\times)v_{eb}^e+C_b^ef_{ib}^b+g_{ib}^e & v_{eb}^e\\
	0_{1\times 3}&0&0\\0_{1\times 3}&0&0
	\end{bmatrix}\\
	=&\begin{bmatrix}
	C_b^e & v_{eb}^e & r_{eb}^e\\0_{1\times 3}&1&0\\0_{1\times 3}&0&1
	\end{bmatrix}\begin{bmatrix}
	\omega_{ib}^b\times & f_{ib}^b & 0_{3\times 1}\\ 0_{1\times 3} &0&0\\ 0_{1\times 3}&0&0
	\end{bmatrix}+\\
	&\begin{bmatrix}
	-\omega_{ie}^e\times & g_{ib}^e-\omega_{ie}^e\times v_{eb}^e & v_{eb}^e+\omega_{ie}^e\times r_{eb}^e\\ 0_{1\times 3} &0&0\\ 0_{1\times 3}&0&0
	\end{bmatrix}\begin{bmatrix}
	C_b^e & v_{eb}^e & r_{eb}^e\\0_{1\times 3}&1&0\\0_{1\times 3}&0&1
	\end{bmatrix}
	\triangleq \mathcal{X}W_1+W_2\mathcal{X}
	\end{aligned}
	\end{equation}
	It is easy to verify that the dynamical equation is group-affine property similar to equation(\ref{proof_invariance}). 
	
	%%%%%%%%%%%%%%%%%%%%%%%%%%%%%%%%%%%%%%%%%%%%%%%%%%%%%%%%%%%%%%%%%%%%%%%%%%%%%
\subsection{Kinematic Equations on transformed ECEF Navigation}
Similar to the auxiliary velocity defined by equation(\ref{auxiliary_velocity}) in the navigation frame, for the inertial-integrated navigation in ECEF frame, a new auxiliary velocity can be defined as
\begin{equation}\label{auxiliary_velocity_ECEF}
	{v}_{ib}^e=v_{eb}^e+\omega_{ie}^e\times r_{eb}^e
\end{equation}

Then, the error state dynamical equation can be manipulated in parallel to the manipulation in section \ref{transformed_INS}, so the similar $SE_2(3)$ based filtering algorithms in ECEF frame are naturally obtained.
	When the system state is defined as
	\begin{equation}\label{new_state_ie_n}
	\mathcal{X}=\begin{bmatrix}
	C_b^e & v_{ib}^e & r_{ib}^e\\
	0_{1\times3} & 1 & 0\\
	0_{1\times 3} & 0& 1
	\end{bmatrix}\in SE_2(3)
	\end{equation}
	where $C_b^e$ is the direction cosine matrix from the body frame to the ECEF frame; $v_{ib}^e$ is the velocity of body frame relative to the ECI frame expressed in the ECEF frame; $r_{ib}^e$ is the position of body frame relative to the ECI frame expressed in the ECEF frame.
	
	Then the dynamic equation of the state $\mathcal{X}$  can be deduced as follows
	\begin{equation}\label{lie_group_state_differential}
	\begin{aligned}
	&\frac{d}{dt}\mathcal{X}=f_{u_t}(\mathcal{X})=\frac{d}{dt}\begin{bmatrix}
	C_b^e & v_{ib}^e & r_{ib}^e\\0_{1\times 3}&1&0\\0_{1\times 3}&0&1
	\end{bmatrix}=\begin{bmatrix}
	\dot{C}_b^e & \dot{v}_{ib}^e & \dot{r}_{ib}^e\\0_{1\times 3}&0&0\\0_{1\times 3}&0&0
	\end{bmatrix}\\
	=&\begin{bmatrix}
	C_b^e(\omega_{ib}^b\times)-(\omega_{ie}^e\times)C_b^e & (-\omega_{ie}^e\times)v_{ib}^e+C_b^ef_{ib}^b+G_{ib}^e & (-\omega_{ie}^e\times)r_{ib}^e+v_{ib}^e\\
	0_{1\times 3}&0&0\\0_{1\times 3}&0&0
	\end{bmatrix}\\
	=&\begin{bmatrix}
	C_b^e & v_{ib}^e & r_{ib}^e\\0_{1\times 3}&1&0\\0_{1\times 3}&0&1
	\end{bmatrix}\begin{bmatrix}
	\omega_{ib}^b\times & f_{ib}^b & 0_{3\times 1}\\ 0_{1\times 3} &0&0\\ 0_{1\times 3}&0&0
	\end{bmatrix}+\begin{bmatrix}
	-\omega_{ie}^e\times & G_{ib}^e & v_{ib}^e\\ 0_{1\times 3} &0&0\\ 0_{1\times 3}&0&0
	\end{bmatrix}\begin{bmatrix}
	C_b^e & v_{ib}^e & r_{ib}^e\\0_{1\times 3}&1&0\\0_{1\times 3}&0&1
	\end{bmatrix}\\
	=&\mathcal{X}W_1+W_2\mathcal{X}
	\end{aligned}
	\end{equation}
	It is easy to verified the dynamical equation (\ref{lie_group_state_differential}) satisfies the group-affine property similar to equation~(\ref{proof_invariance}).
%%%%%%%%%%%%%%%%%%%%%%%%%%%%%%%%%
\section{Exact Discrete Model}
As we associate navigation state $C,v,r$ with an element $\mathcal{X}$ of matrix Lie group $SE_2(3)$, it is easy to conclude that the dynamical equations of the navigation sate can be written on matrix Lie group as
\begin{equation}\label{dynamical_equation}
\frac{d}{d t}\mathcal{X}=f_{u_t}(\mathcal{X})=\mathcal{X}W_1+W_2\mathcal{X}
\end{equation}
where $f_{u_t}(\cdot)$ is proofed to be group affine.
Is worth noting that $W_1$ is only related to the measurement variables which enlighten us whether we can integrate the navigation state related to instrument measurement separately.

Furthermore, the dynamical equations can also be written in the following decomposition form:
\begin{equation}\label{dynamical_equation_three_terms}
\frac{d}{d t}\mathcal{X}_t=W_t\mathcal{X}_t+f(\mathcal{X}_t)+\mathcal{X}_tU_t
\end{equation}
where $f(\cdot)$ is supposed to be group affine.

In particular, for the transformed INS dynamical equations in navigation frame, $W_t$, $U_t$, and $f(\mathcal{X}_t)$ are given as
\begin{equation}\label{WUf_v_in}
W_t=\begin{bmatrix}
-(\omega_{in}^n\times) & G_{in}^n & 0_{3\times 1}\\0_{2\times 3} & 0_{2\times 1} & 0_{2\times 1}
\end{bmatrix}, U_t=\begin{bmatrix}
\omega_{ib}^b\times & f_{ib}^b & 0_{3\times 1}\\
0_{2\times 3} & 0_{2\times 1} & 0_{2\times 1}
\end{bmatrix}, f(\mathcal{X}_t)=\begin{bmatrix}
0_{3\times 3} & 0_{3\times 1} & v_{in}^n\\
0_{2\times 3} & 0_{2\times 1} & 0_{2\times 1}
\end{bmatrix}
\end{equation}

For the transformed INS dynamical equations in ECEF frame, $W_t$, $U_t$, and $f(\mathcal{X}_t)$ are given as
\begin{equation}\label{WUf_v_ib}
W_t=\begin{bmatrix}
-(\omega_{ie}^e\times) & G_{ib}^e & 0_{3\times 1}\\0_{2\times 3} & 0_{2\times 1} & 0_{2\times 1}
\end{bmatrix}, U_t=\begin{bmatrix}
\omega_{ib}^b$  $\times & f_{ib}^b & 0_{3\times 1}\\
0_{2\times 3} & 0_{2\times 1} & 0_{2\times 1}
\end{bmatrix}, f(\mathcal{X}_t)=\begin{bmatrix}
0_{3\times 3} & 0_{3\times 1} & v_{ib}^e\\
0_{2\times 3} & 0_{2\times 1} & 0_{2\times 1}
\end{bmatrix}
\end{equation}

The solution $\mathcal{X}_t$ at arbitrary $t$ of equation (\ref{dynamical_equation_three_terms}) can be written as a function of the initial value $\mathcal{X}_0$ in compact and intrinsic form through group multiplication~\cite{barrau:hal-01671724}:
\begin{equation}\label{decomposition_initial}
\mathcal{X}_t=\Gamma_t\Phi_t(\mathcal{X}_0)\Upsilon_t
\end{equation}
where $\mathcal{X}_t$ is a global navigation state, $\Gamma_t,\Upsilon_t\in SE_2(3)$ are solution to differential equations involving only $W_t$, $U_t$, respectively, and where $\Phi_t(\cdot)$ is a function that only depends on $t$. 

Solving the corresponding equations yields
\begin{equation}\label{phi_gamma_upsilon}
\begin{aligned}
&\Phi_t:\begin{bmatrix}
C & v & r\\
0& 1&0\\
0&0&1
\end{bmatrix}\mapsto \begin{bmatrix}
C & v & r+tv\\
0& 1&0\\
0&0&1
\end{bmatrix}=\mathcal{X}+tf(\mathcal{X}), \\
&\Gamma_t=\begin{bmatrix}
\Gamma_t^C & \Gamma_t^v & \Gamma_t^r\\
0& 1&0\\
0&0&1
\end{bmatrix}, \Upsilon_t=\begin{bmatrix}
\Delta C & \Delta v & \Delta r\\
0& 1&0\\
0&0&1
\end{bmatrix}
\end{aligned}
\end{equation}
where $\Gamma_t$ is the solution of
\begin{equation}\label{Gamma_differential}
\frac{d}{dt}\Gamma_t=W_t\Gamma_t+f(\Gamma_t)
\end{equation}
 and $\Upsilon_t$ is the solution of
 \begin{equation}\label{Upsilon_differential}
 \frac{d}{dt}\Upsilon_t=\Upsilon_t U_t+f(\Upsilon_t)
 \end{equation}
 with initial condition $\Gamma_0=\Upsilon_0=I_{5\times 5} $.

Solving the corresponding equations in the particular case of equations~(\ref{differential_invariant}) on $SE_2(3)$ with values given by (\ref{WUf_v_in}), quantities in $\Gamma_t$ are defined by
\begin{equation}\label{Gamma_1}
\Gamma_0^C=I,\frac{d}{dt}\Gamma_t^C=-\omega_{in}^n\times \Gamma_t^C, \Gamma_0^v=0,\frac{d}{dt}\Gamma_t^v=G_{in}^n-\omega_{in}^n\times \Gamma_t^v,
\Gamma_0^r=0,\frac{d}{dt}\Gamma_t^r=\Gamma_t^v-\omega_{in}^n\times \Gamma_t^r
\end{equation}
and quantities in $\Upsilon_t$ are defined by
\begin{equation}\label{Upsilon_1}
\Delta C_{0}=I,\frac{d}{dt}\Delta C=\Delta C(\omega_{ib}^b\times ), \Delta v_{0}=0, \frac{d}{dt}\Delta v=\Delta Cf_{ib}^b, \Delta r_{0}=0, \frac{d}{dt}\Delta r=\Delta v
\end{equation}

Solving the corresponding equations in the particular case of equations~(\ref{lie_group_state_differential}) on $SE_2(3)$ with values given by (\ref{WUf_v_ib}), only the quantities in $\Gamma_t$ are different and defined by
\begin{equation}\label{Gamma_2}
\Gamma_0^C=I,\frac{d}{dt}\Gamma_t^C=-\omega_{ie}^e\times \Gamma_t^C, \Gamma_0^v=0,\frac{d}{dt}\Gamma_t^v=G_{ib}^e-\omega_{ie}^e\times \Gamma_t^v,
\Gamma_0^r=0,\frac{d}{dt}\Gamma_t^r=\Gamma_t^v-\omega_{ie}^e\times \Gamma_t^r
\end{equation}

It can easily be verified that $\Phi_t(\cdot)$ is a group automorphism, which means it is invertible and satisfies 
\begin{equation}\label{automorphism_Phi}
\Phi_t(\mathcal{X}^1\mathcal{X}^2)=\Phi_t(\mathcal{X}^1)\Phi_t(\mathcal{X}^2),\Phi_t(\mathcal{X}^{-1})=\Phi_t(\mathcal{X})^{-1}
\end{equation}

For the $\Phi_t(\cdot)$ defined in equation~(\ref{phi_gamma_upsilon}), it owns the log-linearity property, that is \begin{equation}\label{log_linearity_property}
\Phi_t(\exp_G(\Lambda(\xi)))=\exp_G(\Lambda(F\xi)), F:=F_{ t}:=\begin{bmatrix}
I & 0& 0\\
0& I & 0\\
0& tI& I
\end{bmatrix}
\end{equation}

Combine with the two properties, we can get
\begin{equation}\label{automorphism_log_linearity}
\Phi_t(\mathcal{X}\exp_G(\Lambda(\xi)))=\Phi_t(\mathcal{X})\exp_G(\Lambda(F\xi))
\end{equation}

In addition, $\Gamma_t$ is a global increment and $\Upsilon_t$ is a local increment.
%As stated in literature~\cite{forster2016manifold}, the quantities $\Delta C$, $\Delta v$, and $\Delta r$ are referred to as the Delta preintegrated measurements and are based solely on the inertial measurements and do not depend on the initial state $\mathcal{X}_0$. The sophisticated calculation of the preintegrated measurement can be found in literature~\cite{brossard2020associating}.
Therefore, the navigation state with initial state $(C_0, v_{ib,0}^e, r_{ib,0}^e)$ can be written separately as 
\begin{equation}\label{vibe_discrete}
\begin{aligned}
C_{b,t}^e&=\Gamma_t^C C_0 \Delta C\\
v_{ib,t}^e&=\Gamma_t^v+\Gamma_t^C(C_0\Delta v+v_{ib,0}^e)\\
r_{ib,t}^e&=\Gamma_t^r+\Gamma_t^C(C_0\Delta r+r_{ib,0}^e+tv_{ib,0}^e)
\end{aligned}
\end{equation}

The navigation state with initial state $(C_0, v_{in,0}^n, r_{in,0}^n)$ can be written separately as 
\begin{equation}\label{vinn_discrete}
\begin{aligned}
C_{b,t}^n&=\Gamma_t^C C_0 \Delta C\\
v_{in,t}^n&=\Gamma_t^v+\Gamma_t^C(C_0\Delta v+v_{in,0}^n)\\
r_{in,t}^n&=\Gamma_t^r+\Gamma_t^C(C_0\Delta r+r_{in,0}^n+tv_{in,0}^n)
\end{aligned}
\end{equation}

It is worth noting that no approximation is made. Meanwhile, the time $t$ here refers to any time length as the time differential of each element of the navigation state is equivalent to its kinematic equation given in Section~\ref{sectionIII}. We will see how to discretize the differential equations of local increment and global increment, and then update them iteratively in the following.
%%%%%%%%%%%%%%%%%%%%%%%%%%%%%%%%%%%%%%%%%%%%
\subsection{Calculation for global increment}\label{global_increment_section}
On the one hand, we now focus on the numerical integration for global increment defined in equation~(\ref{Gamma_1}). 
It is reasonable to assume a piecewise constant model as the term $\omega_{in}^n$ defined in equation (\ref{omega_ie_n}) changes over time at a very slow speed. 
By leveraging the exponential mapping of $SO(3)$, the term related to attitude can be calculated
\begin{equation}\label{Gamma_Rotation}
\Gamma_{ij}^C=\exp_G((-(\omega_{in}^n)_{t_i}\Delta t_{ij})\times)
\end{equation}

For $\Gamma_t^v$ and $\Gamma_t^r$, their differential equations can be reformulated as the following continuous time linear system:
\begin{equation}\label{continuous_time_linear_system}
\begin{bmatrix}
\dot{\Gamma}_t^v\\ \dot{\Gamma}_t^r
\end{bmatrix}=\begin{bmatrix}
-\omega_{in}^n\times  & 0_{3\times3}\\
I_{3\times 3} & -\omega_{in}^n\times 
\end{bmatrix}\begin{bmatrix}
\Gamma_t^v \\ \Gamma_t^r
\end{bmatrix}+\begin{bmatrix}
G_{in}^n\\0_{3\times 1}
\end{bmatrix}
\end{equation}
This equation is essentially a time varying linear system.
Assume that the value of $\omega_{in}^n$ is fixed within small interval as the change of $\omega_{in}^n$ due to velocity and position is quite small, the state transition matrix is therefore a constant matrix in the interval.
Subsequently, the system in ($\ref{continuous_time_linear_system}$) is a linear-time-invariant system during the interval $\Delta t$.
Therefore, numerical integration schemes such as Euler integration are adopted for the discretization.

For the quantity $\Gamma_t^v$, it can be discretized as
\begin{equation}\label{Gamma_t_v_discretize}
\Gamma_{t_j}^v=\Gamma_{t_{i}}^v+(G_{in}^n-\omega_{in}^n\times \Gamma_{t}^v)_{t_{i}}\Delta t_{ij}
\end{equation}
where $\Delta t_{ij}=t_{j}-t_{i}$ is the time interval.

For the quantity $\Gamma_t^r$, it can be discretized as
\begin{equation}\label{Gamma_t_r_discretize}
\Gamma_{t_j}^r=\Gamma_{t_{i}}^r+(\Gamma_t^v-\omega_{in}^n\times \Gamma_t^r)_{t_{i}}\Delta t_{ij}
\end{equation}

On the other hand, as concerns the global increment defined in equation~(\ref{Gamma_2}), the earth rotation vector expressed in ECEF frame is constant and the continuous time linear system related to $\Gamma_t^v$ and $\Gamma_t^r$ is linear time invariant.
For the component of rotation, the solution can be given as
 \begin{equation}\label{Gamma_Rotation_ecef}
 \Gamma_{ij}^C=\exp_G((-(\omega_{ie}^e)_{t_i}\Delta t_{ij})\times)
 \end{equation}
 
The differential equations for global increment defined in equation~(\ref{Gamma_2}) can be formulated in vector form:
\begin{equation}\label{global_increment_vector_form}
\frac{d}{dt}\begin{bmatrix}
\Gamma_{t}^v\\
\Gamma_{t}^r
\end{bmatrix}=\begin{bmatrix}
 -\omega_{ie}^e\times & 0_{3\times 3}\\
        I_{3\times 3}           & -\omega_{ie}^e\times
\end{bmatrix}\begin{bmatrix}
\Gamma_{t}^v\\
\Gamma_{t}^r
\end{bmatrix}+\begin{bmatrix}
G_{ib}^e\\
0_{3\times 1}
\end{bmatrix}
\end{equation}
It is obvious that this is a first-order linear differential equation. The integration over the interval $[t_i,t_j]$ in such case will result in
\begin{equation}\label{general_solution}
\begin{bmatrix}
\Gamma_{t_j}^v\\
\Gamma_{t_j}^r
\end{bmatrix}=\Psi(t_j,t_i)\begin{bmatrix}
\Gamma_{t_i}^v\\
\Gamma_{t_i}^r
\end{bmatrix}+\int_{t_i}^{t_j}\Psi(t_j,s)\begin{bmatrix}
G_{ib}^e\\
0_{3\times 1}
\end{bmatrix}ds
\end{equation}
where $\Psi(t_j,s)$ is known as the transition function and the Taylor expansion of this transition matrix is
\begin{equation}\label{transition_matrix}
\begin{aligned}
\Psi(t_j,s)=&\exp_G\left( \begin{bmatrix}
 -\omega_{ie}^e\times & 0_{3\times 3}\\
       I_{3\times 3}           & -\omega_{ie}^e\times
\end{bmatrix}(t_j-s) \right)\\
=&\begin{bmatrix}
 Exp(-\omega_{ie}^e(t_j-s)) &0_{3\times 3}\\
(t_j-s)Exp(-\omega_{ie}^e(t_j-s)) &Exp(-\omega_{ie}^e(t_j-s)) 
\end{bmatrix}
\end{aligned}
\end{equation}

%Therefore, the integration of global increment in vector form is given as
%\begin{equation}\label{general_solution1}
%\begin{bmatrix}
%{\Gamma}_{t_j}^C \\
%\Gamma_{t_j)}^v\\
%\Gamma_{t_j}^r
%\end{bmatrix}=\Psi(t_j,t_i)\begin{bmatrix}
%{\Gamma}_{t_i}^C \\
%\Gamma_{t_i}^v\\
%\Gamma_{t_i}^r
%\end{bmatrix}+\int_{t_i}^{t_j}\Psi(t_j,s)\begin{bmatrix}
%0_{3\times 1} \\
%G_{ib}^e\\
%0_{3\times 1}
%\end{bmatrix}ds
%\end{equation}

The recursive formula for the global increment can be written as
\begin{equation}\label{global_increment_recursive}
\begin{aligned}
\Gamma_{j}&=\Gamma_{ij}\Phi_{ij}(\Gamma_{i})\\
\Rightarrow \begin{bmatrix}
\Gamma_{j}^C & \Gamma_{j}^v & \Gamma_{j}^r\\
0_{1\times 3} & 1 & 0\\
0_{1\times 3} &0 & 1
\end{bmatrix}&=\begin{bmatrix}
\Gamma_{ij}^C & \Gamma_{ij}^v & \Gamma_{ij}^r\\
0_{1\times 3} & 1 & 0\\
0_{1\times 3} &0 & 1
\end{bmatrix}\begin{bmatrix}
\Gamma_{i}^C & \Gamma_{i}^v & \Gamma_{i}^r+\Delta t_{ij}\Gamma_{i}^v\\
0_{1\times 3} & 1 & 0\\
0_{1\times 3} &0 & 1
\end{bmatrix}\\
\Rightarrow \begin{bmatrix}
\Gamma_{j}^C \\ \Gamma_{j}^v \\ \Gamma_{j}^r
\end{bmatrix}&=\begin{bmatrix}
\Gamma_{ij}^C & 0_{3\times 3} & 0_{3\times 3}\\
0_{3\times 3}& \Gamma_{ij}^C &0_{3\times 3}\\
0_{3\times 3}&\Delta t_{ij}\Gamma_{ij}^C &\Gamma_{ij}^C 
\end{bmatrix}\begin{bmatrix}
\Gamma_{i}^C \\ \Gamma_{i}^v \\ \Gamma_{i}^r
\end{bmatrix}+\begin{bmatrix}
0_{3\times 3} \\ \Gamma_{ij}^v \\ \Gamma_{ij}^r
\end{bmatrix}
\end{aligned}
\end{equation}
where 
\begin{equation}\label{Gamma_ij_C_v_r}
\begin{aligned}
\Gamma_{ij}^C&=Exp(-\omega_{ie}^e(t_j-t_i))=Exp(-\omega_{ie}^e \Delta t_{ij})\\
\Gamma_{ij}^v&=\int_{t_i}^{t_j}  Exp(-\omega_{ie}^e(t_j-s)) G_{ib}^e ds=\int_{0}^{\Delta t_{ij}}Exp(-\omega_{ie}^e u)G_{ib}^edu =\Gamma_{1}(-\omega_{ie}^e\Delta t_{ij})\Delta t_{ij} G_{ib}^e\\ 
\Gamma_{ij}^r&=\int_{t_i}^{t_j} (t_j-s)Exp(-\omega_{ie}^e(t_j-s)) G_{ib}^e ds=\int_{0}^{\Delta t_{ij}}uExp(-\omega_{ie}^e u)G_{ib}^edu\\
&=\Gamma_{2}(-\omega_{ie}^e\Delta t_{ij})\Delta t_{ij}^2 G_{ib}^e
\end{aligned}
\end{equation}

Therefore, exact closed-form expression of $\Gamma_t$ can be obtained and is the same with that of~\cite{brossard2021associating}.

Going back to the differential equations of the global increment in transformed NED navigation frame, equation (\ref{Gamma_1}) can be reformulated as vector form, that is
\begin{equation}\label{global_increment_vector_form_transformed_ned}
\frac{d}{dt}\begin{bmatrix}
\Gamma_{t}^v\\
\Gamma_{t}^r
\end{bmatrix}=\begin{bmatrix}
 -\omega_{in}^n\times & 0_{3\times 3}\\
       I_{3\times 3}           & -\omega_{in}^n\times
\end{bmatrix}\begin{bmatrix}
\Gamma_{t}^v\\
\Gamma_{t}^r
\end{bmatrix}+\begin{bmatrix}
G_{ib}^n\\
0_{3\times 1}
\end{bmatrix}
\end{equation}
The above equation is concerned with $\omega_{in}^n$, which is a function of velocity and position and changes slowly. 
Therefore, the time variant system in equation (\ref{global_increment_vector_form_transformed_ned}) can be treated as a switched linear system between two consecutive time instances $[t_i,t_{i+1}]$~\cite{li2005switched}. Similar to the global increment calculated in transformed ECEF frame,  the global increment between two consecutive time instances in transformed NED navigation frame can be calculated as
\begin{equation}\label{Gamma_ij_C_v_r_transformed_ned}
\begin{aligned}
\Gamma_{i,i+1}^C&=Exp(-\omega_{in}^n(t_{i+1}-t_i))=Exp(-(\omega_{in}^n)_{t_i} \Delta t_{i,i+1})\\
\Gamma_{i,i+1}^v&=\int_{t_i}^{t_{i+1}}  Exp(-\omega_{in}^n(t_{i+1}-s)) G_{ib}^n ds=\int_{0}^{\Delta t_{i,{i+1}}}Exp(-\omega_{in}^n u)G_{ib}^ndu\\ &=\Gamma_{1}(-(\omega_{in}^n)_{t_i}\Delta t_{i,{i+1}})\Delta t_{i,{i+1}} G_{ib}^n\\
\Gamma_{i,i+1}^r&=\int_{t_i}^{t_{i+1}} (t_{i+1}-s)Exp(-\omega_{in}^n(t_{i+1}-s)) G_{ib}^n ds=\int_{0}^{\Delta t_{i,{i+1}}}uExp(-\omega_{in}^n u)G_{ib}^ndu\\
&=\Gamma_{2}(-(\omega_{in}^n)_{t_i}\Delta t_{i,{i+1}})\Delta t_{i,{i+1}}^2 G_{ib}^n
\end{aligned}
\end{equation}
where $\omega_{in}^n$ and $G_{ib}^n$ are considered as constants during the update interval and take the value at instance $t_i$.

In order to obtain the global increment between arbitrary time instance $t_i$ and $t_j$, the recursive formula for $\Gamma_{ij}$ is given as
\begin{equation}\label{iterative_update_gamma_ij}
\Gamma_{ij}=\Gamma_{j-1,j}\Phi_{j-1,j}(\Gamma_{i,j-1})
\end{equation}
%%%%%%%%%%%%%%%%%%%%%%%%%%%%%%%%%%%%%%%%%%%%%%
\subsection{Calculation for Local increment}
There are several sophisticated schemes for calculation of the IMU local increment $\Upsilon_t$ for one time step~\cite{ brossard2021associating}. There two close-form expressions for the local increment based on two basic assumptions, namely piecewise constant global acceleration~\cite{forster2016manifold} and piecewise constant IMU measurements~\cite{eckenhoff2019closed}.
In fact, the quaternion-based integration model proposed by \cite{eckenhoff2019closed} is equivalent to accurate IMU preintegration using switched linear systems proposed by \cite{henawy2019accurate,henawy2021accurate} which also assumes that the measurements are constant in local frame instead of the world frame.
Particularly, the earth's rotation is considered for both two assumptions using extended pose representation in \cite{brossard2021associating}. 
%The improvement brought by considering the earth's rotation is evaluated by ~\cite{tang2021exploring}, whereas attitude is still represented by quaternion.

The state estimation at $t+\Delta t$ can be obtained by the Euler integration of kinematic equation (\ref{Upsilon_differential}) from $t$ to $t+\Delta t$ as
\begin{equation}\label{discrete_Euler_integration}
\begin{aligned}
\Delta R(t+\Delta t)&=\Delta R(t)Exp(\int_{t}^{t+\Delta t}\omega_{ib}^b(\tau)d\tau)\\
\Delta v(t+\Delta t)&=\Delta v(t)+\int_{t}^{t+\Delta t} \Delta R(\tau) f_{ib}^b(\tau)d\tau\\
\Delta r(t+\Delta t)&=\Delta r(t)+\int_{t}^{t+\Delta t}\Delta v(\tau)d\tau +\int\int_{t}^{t+\Delta t}\Delta R(\tau)f_{ib}^b(\tau)d\tau^2
\end{aligned}
\end{equation}
In general, there is no closed form solution for the above equation as the measurements are time varying aceeleration and angular velocity. 
It is worth noting that the preintegration measurements is referred to the local increment $\Upsilon_{ij}$ between time interval $t_i$ and $t_j$, instead of the quantity $\Upsilon_i$ at time $t_i$.

By assuming that both the linear acceleration in the inertial frame $\Delta R f_{ib}^b$ and the angular velocity in the body frame $\omega_{ib}^b$ are constant between two successive IMU measurement, the discrete kinematic model can be expressed as
\begin{equation}\label{discrete_Euler_integration_increment}
\begin{aligned}
\Delta R(t+\Delta t)&=\Delta R(t)Exp(\omega_{ib}^b\Delta t)\\
\Delta v(t+\Delta t)&=\Delta v(t)+ \Delta R(t) f_{ib}^b \Delta t\\
\Delta r(t+\Delta t)&=\Delta r(t)+\Delta v\Delta t +\frac{1}{2}\Delta R(t)f_{ib}^b\Delta t^2
\end{aligned}
\end{equation}
It is worth noting that $\Delta r(t+\Delta t)=\Delta r(t)+\frac{1}{2}\left(\Delta v+\Delta v(t+\Delta t)\right)\Delta t$, which is equivalent to the midpoint integration.

The IMU preintegration between time instance $k=i$ and time instance $k=j$ can be obtained by the following equation
\begin{equation}\label{preintegration_i_j}
\begin{aligned}
\Delta R(j)&=\Delta R(i)F(i,j)\\
\Delta v(j)&=\Delta v(i)+\Delta R(i)\mu_1(i,j)\\
\Delta r(j)&=\Delta r(i)+\Delta v(i)\sum_{k=i}^{j-1}\Delta t+\Delta R(i)\zeta_1(i,j)
\end{aligned}
\end{equation}
where the matrix $F(i,j)$ is computed as
\begin{equation}\label{F_i_j}
F(i,j)=\prod_{k=i}^{j-1}Exp(\omega_{ib}^b\Delta t)
\end{equation}
and the vectors $\mu_1(i,j)$ and $\zeta_1(i,j)$ are computed as
\begin{equation}\label{mu_1}
\mu_1(i,j)=\sum_{k=i}^{j-1}F(i,k)f_{ib}^b(k)\Delta t
\end{equation}
\begin{equation}\label{zeta_1}
\zeta_1(i,j)=\sum_{k=i}^{j-1}\left(\frac{1}{2}F(i,k)f_{ib}^b(k)\Delta t^2 \right)
\end{equation}

But it is more intuitive to assume the local acceleration to be constant instead of the global acceleration.
By assuming that both the linear acceleration and the angular velocity are constant in body frame between two successive IMU measurement, the attitude can be updated as
\begin{equation}\label{attitude_update_zero_order}
\Delta R(t+\Delta t)=\Delta R(t)Exp(\omega_{ib}^b\Delta t)
\end{equation}
The velocity can be updated by integrating the dynamics of velocity which involves the integral of the exponential mapping:
\begin{equation}\label{velocity_update_zero_order}
\begin{aligned}
\Delta v(t+\Delta t)&=\Delta v(t)+\int_{t}^{t+\Delta t} \Delta R(\tau) f_{ib}^b(\tau)d\tau\\
&=\Delta v(t)+\Delta R(t)\left( \int_{t}^{t+\Delta t} Exp(\omega_{ib}^b\tau)d\tau \right) f_{ib}^b\\
&=\Delta v(t)+\Delta R(t)\Gamma_1(\omega_{ib}^b(k)\Delta t)\Delta t f_{ib}^b
\end{aligned}
\end{equation}
Likewise, the position can be updated by integrating the dynamics of position which involves computing the double integral:
\begin{equation}\label{postion_update_zero_order}
\begin{aligned}
\Delta r(t+\Delta t)&=\Delta r(t)+\int_{t}^{t+\Delta t}\Delta v(\tau)d\tau +\int\int_{t}^{t+\Delta t}\Delta R(\tau)f_{ib}^b(\tau)d\tau^2\\
&=\Delta r(t)+\Delta v(t)\Delta t +\Delta R(t)\left( \int_{t}^{t+\Delta t}\int_{t}^{t+ \tau} Exp(\omega_{ib}^b\tau_1)d\tau_1 d\tau  \right) f_{ib}^b\\
&=\Delta r(t)+\Delta v(t)\Delta t +\Delta R(t)\Gamma_2(\omega_{ib}^b(t)\Delta t) \Delta t^2f_{ib}^b
\end{aligned}
\end{equation}
The above discrete dynamics of attitude, velocity, and position are an exact integration of the differential equation (\ref{Upsilon_differential}) by assuming a zero-order hold on the IMU measurements.
Therefore, the IMU preintegration theory model under the zero-order hold are given as
\begin{equation}\label{accurate_IMU_preintegration}
\begin{aligned}
\Delta R(j)&=\Delta R(i)F(i,j)\\
\Delta v(j)&=\Delta v(i)+\Delta R(i)\mu_2(i,j)\\
\Delta r(j)&=\Delta r(i)+\Delta v(i)\sum_{k=i}^{j-1}\Delta t+\Delta R(i)\zeta_2(i,j)
\end{aligned}
\end{equation}
where the vectors $\mu_2(i,j)$ and $\zeta_2(i,j)$ are computed as
\begin{equation}\label{mu_2}
\mu_2(i,j)=\sum_{k=i}^{j-1}F(i,k)\Gamma_1(\omega_{ib}^b(k)\Delta t)f_{ib}^b(k)\Delta t
\end{equation}
\begin{equation}\label{zeta_2}
\zeta_2(i,j)=\sum_{k=i}^{j-1}\left(F(i,k)\Gamma_2(\omega_{ib}^b(k)\Delta t)f_{ib}^b(k)\Delta t^2 \right)
\end{equation}

Comparing the results obtained above with the accurate IMU preintegration model proposed by ~\cite{henawy2021accurate}, it is easy to find that the differences between them are in terms of the gravity acceleration and the Earth rotation rate.

However, most algorithms assume fixed axis rotation of the moving frame in the renewal period, this will inevitably introduce non-commutative error if the rotation is not a fixed axis rotation.
% and the coning error optimization algorithm are used to improve the solution accuracy.
Coning, sculling, and scrolling compensated multi-sample algorithms can also be used to improve the solution accuracy.
In the high dynamic environment, polynomial iteration base equivalent rotation vector is adopted as the accurate numerical solution for the local increment which assumes the measurements in local frame is polynomial instead of constant~\cite{gongmin2017lectures}.
In the vibrating environment, the coning error will be introduced under the condition of coning motion and multi-sample optimized coning compensation algorithms can be adopted~\cite{gongmin2017lectures}.
Here, we just take the two-sample error compensation algorithm as an example, and incorporate the multi-sample error compensation algorithms into the preintegration theory model.

In fact, the kinematic equation (\ref{Upsilon_differential}) can be treated as the attitude, velocity, and position expressed in the ECI frame, and it also can be expressed equivalently as
\begin{equation}\label{c_b_i}
\begin{aligned}
\dot{R}_b^i&=R_b^i(\omega_{ib}^b\times)\\
\dot{v}_{ib}^i&=R_b^if_{ib}^b\\
\dot{r}_{ib}^i&=v_{ib}^i
\end{aligned}
\end{equation}

Therefore, the update algorithm for attitude can be expressed as
\begin{equation}\label{attitude_update}
R_{b(m)}^{i}=R_{b(m-1)}^iR_{b(m)}^{b(m-1)}=R_{b(m-n)}^iR_{b(m-1)}^{b(m-n)}R_{b(m)}^{b(m-1)}
\end{equation}
where $m$ represents the time instance $t_m$; $R_{b(m)}^{b(m-1)}$ represents the attitude change of the body frame from time $t_{m-1}$ to time $t_m$ with the ECI frame as reference and it can be calculated by the angular velocity measurement provided by gyroscope. When we represent the attitude by equivalent rotation vector, the two-sample coning error compensation algorithm is adopted under the assumption that the output of angular velocity is linear in time, that is~\cite{gongmin2017lectures}
\begin{equation}\label{rotation_vector}
R_{b(m)}^{b(m-1)}=\Gamma_0(\phi_{ib(m)}^b), \phi_{ib(m)}^b=(\Delta \theta_{m1}+\Delta \theta_{m2})+\frac{2}{3}\Delta \theta_{m1}\times\Delta \theta_{m2}
\end{equation}
where $\Delta \theta_{m1}$ and $\Delta\theta_{m2}$ are the incremental-angle vector over the $2-th$ minor interval which can be expressed as $\Delta\theta_{m1}=\int_{t_{m-1}}^{t_{m-1}+\frac{1}{2}\Delta t}\omega_{ib}^{b}(\tau)d\tau$ and $\Delta \theta_{m2}=\int_{t_{m-1}+\frac{1}{2}\Delta t}^{t_{m}}\omega_{ib}^{b}(\tau)d\tau$ respectively. Of course, there are lots of coning error correction algorithms for the attitude update.
\begin{remark}
	Although there are lots of error compensation algorithm, the two-sample algorithms are enough for the MEMS-grade IMU which improves accuracy of the attitude but does not increase too much cost of calculation when compared with the single-sample algorithm.
\end{remark}

The attitude preintegration measurement is define as
\begin{equation}\label{attitude_preintegration_measurement}
R_{b(m)}^{b(m-n)}=\mathcal{F}(m-n+1,m)=\prod_{k=m-n+1}^{m}\Gamma_0(\phi_{ib(k)}^b)\Rightarrow \mathcal{F}(i,j)=\prod_{k=i}^{j-1}\Gamma_0(\phi_{ib(k)}^b)
\end{equation}

The integration of the velocity is given as
\begin{equation}\label{velocity_preintegration_measurement}
v_{ib(m)}^i-v_{ib(m-1)}^i=\int_{t_{m-1}}^{t_m} R_b^if_{ib}^b d\tau =\int_{t_{m-1}}^{t_m} R_{b(m-1)}^iR_{b(\tau)}^{b(m-1)}f_{ib}^{b(\tau)} d\tau=R_{b(m-1)}^i\int_{t_{m-1}}^{t_m} R_{b(\tau)}^{b(m-1)}f_{ib}^{b(\tau)} d\tau
\end{equation}
The equivalent rotation vector corresponding to $R_{b(\tau)}^{b(m-1)}$ is $\phi_{ib}^b(\tau,t_m)$.
When it is assumed to be small quantity, it can be approximated as
\begin{equation}\label{approximation}
R_{b(\tau)}^{b(m-1)}\approx I+\phi_{ib}^b(\tau,t_m)\times 
\end{equation}

Substituting equation (\ref{approximation}) into equation (\ref{velocity_preintegration_measurement}), we can get
\begin{equation}\label{velocity_preintegration_measurement_1}
\begin{aligned}
v_{ib(m)}^i-v_{ib(m-1)}^i&\approx R_{b(m-1)}^i\int_{t_{m-1}}^{t_m} \left(I+\phi_{ib}^b(\tau,t_m)\times \right)f_{ib}^{b(\tau)} d\tau\\
&=R_{b(m-1)}^i\int_{t_{m-1}}^{t_m} f_{ib}^{b(\tau)} d\tau+R_{b(m-1)}^i\int_{t_{m-1}}^{t_m} \phi_{ib}^b(\tau,t_m)\times f_{ib}^{b(\tau)} d\tau\\
&=R_{b(m-1)}^i\Delta v_m+R_{b(m-1)}^i\int_{t_{m-1}}^{t_m} \phi_{ib}^b(\tau,t_m)\times f_{ib}^{b(\tau)} d\tau\\
&=R_{b(m-n)}^iR_{b(m-1)}^{b(m-n)}\delta v_{(m-1)}, \delta v_{(m-1)}\triangleq \Delta v_m+\int_{t_{m-1}}^{t_m} \phi_{ib}^b(\tau,t_m)\times f_{ib}^{b(\tau)} d\tau
\end{aligned}
\end{equation}
where $\Delta v_m=v_{sf}^b(t_m,t_{m-1})=\int_{t_{m-1}}^{t_m} f_{ib}^{b(\tau)} d\tau$ is the velocity increment of the accelerometer sampling.
If the two-sample error compensated algorithms are adopted, the integration $\int_{t_{m-1}}^{t_m} \phi_{ib}^b(\tau,t_m)\times f_{ib}^{b(\tau)} d\tau$ can be calculated as~\cite{gongmin2017lectures}
\begin{equation}\label{integration}
\int_{t_{m-1}}^{t_m} \phi_{ib}^b(\tau,t_m)\times f_{ib}^{b(\tau)} d\tau=\frac{1}{2}\Delta \theta_{m}\times \Delta v_m+\frac{2}{3}(\Delta\theta_{m1}\times\Delta v_{m_2}+\Delta v_{m_1}\times \Delta\theta_{m_2})
\end{equation}
where $\Delta \theta_m=\theta_{ib}^b(t_m,t_{m-1})=\int_{t_{m-1}}^{t_m} \omega_{ib}^{b(\tau)} d\tau$ is the angular increment of the gyroscope sampling, $\Delta v_{m_1}$ and $\Delta v_{m_2}$ are defined as $\Delta v_{m_1}=\int_{t_{m-1}}^{t_{m-1}+\frac{1}{2}\Delta t}f_{ib}^{b(\tau)}d\tau$ and $\Delta v_{m_2}=\int_{t_{m-1}+\frac{1}{2}\Delta t}^{t_{m}}f_{ib}^{b(\tau)}d\tau$, respectively.

It can be shown that equation (\ref{velocity_preintegration_measurement_1}) is equivalent to
\begin{equation}\label{velocity_preintegration_measurement_2}
v_{ib(m)}^i=v_{ib(m-n)}^i+R_{b(m-n)}^i\sum_{k=m-n}^{m-1}R_{b(k)}^{b(m-n)}\delta v_{k}
\end{equation}

The velocity preintegration measurement is defined as
\begin{equation}\label{mu_3}
R_i^{b(m-n)}(v_{ib(m)}^i-v_{ib(m-n)}^i)=\sum_{k=m-n}^{m-1}R_{b(k)}^{b(m-n)}\delta v_{(k)} \Rightarrow \mu_3(i,j)=\sum_{k=i}^{j-1}\mathcal{F}(i,k)\delta v_{(k)}
\end{equation}

With the results of velocity update, the position update is given as
%\begin{equation}\label{position_integration11}
%\begin{aligned}
%r_{ib(m)}^i-r_{ib(m-1)}^i\approx v_{ib(m)}^i\Delta t \overset{(\ref{velocity_preintegration_measurement_2})}{=}v_{ib(m-n)}^i\Delta t+R_{b(m-n)}^i\sum_{k=m-n}^{m-1}R_{b(k)}^{b(m-n)}\delta v_{k}\Delta t
%\end{aligned}
%\end{equation}
%\begin{equation}\label{position_integration12}
%\begin{aligned}
%r_{ib(m)}^i-r_{ib(m-1)}^i\approx v_{ib(m-1)}^i\Delta t \overset{(\ref{velocity_preintegration_measurement_2})}{=}v_{ib(m-n)}^i\Delta t+R_{b(m-n)}^i\sum_{k=m-n}^{m-2}R_{b(k)}^{b(m-n)}\delta v_{k}\Delta t
%\end{aligned}
%\end{equation}
\begin{equation}\label{position_integration13}
\begin{aligned}
&r_{ib(m)}^i-r_{ib(m-1)}^i\approx 0.5(v_{ib(m-1)}^i+v_{ib(m)}^i)\Delta t\\
\overset{(\ref{velocity_preintegration_measurement_1})}{=}&v_{ib(m-1)}^i\Delta t
%+R_{b(m-n)}^i\sum_{k=m-n}^{m-2}R_{b(k)}^{b(m-n)}\delta v_{k}\Delta t
+0.5R_{b(m-n)}^iR_{b(m-1)}^{b(m-n)}\delta v_{(m-1)}\Delta t
\end{aligned}
\end{equation}

It can be shown that equation (\ref{position_integration13}) is equivalent to
\begin{equation}\label{position_integration_1}
\begin{aligned}
R_i^{b(m-n)}(r_{ib(m)}^i-r_{ib(m-n)}^i-\sum_{k=m-n}^{m-1} v_{ib(m-n)}^i\Delta t)=\sum_{k=m-n}^{m-1}
\left(0.5R_{b(k)}^{b(m-n)}\delta v_{k}\Delta t\right)      
\end{aligned}
\end{equation}

The position preintegration measurement is defined as
\begin{equation}\label{zeta_3}
\zeta_3(i,j)=\sum_{k=i}^{j-1}\left(0.5\mathcal{F}(i,k)\delta v_{(k)} \Delta t \right)
\end{equation}

Recently, the Gaussian process regression, which is a non-parametric probabilistic interpolation, is also used to construct a Gaussian Process Preintegration based on continuous representation of inertial measurement~\cite{le2020gaussian, le2021continuous}. 

We finally obtain the local increment by a recursive formula
\begin{equation}\label{local_increment_recursive}
\begin{aligned}
\Upsilon_j&=\Phi_{ij}(\Upsilon_i)\Upsilon_{ij}\\
\Rightarrow \begin{bmatrix}
\Delta R_j & \Delta v_j & \Delta r_j\\
0_{1\times 3}&1&0\\
0_{1\times 3}&0&1
\end{bmatrix}&=\begin{bmatrix}
\Delta R_i & \Delta v_i & \Delta r_i+\Delta t_{ij}\Delta v_i\\
0_{1\times 3}&1&0\\
0_{1\times 3}&0&1
\end{bmatrix}
\begin{bmatrix}
F({i,j}) & \mu({i,j}) & \zeta({i,j})\\
0_{1\times 3}&1&0\\
0_{1\times 3}&0&1
\end{bmatrix}
\end{aligned}
\end{equation}
where $\Upsilon_{ij}$ is the preintegration over the interval $[t_i,t_j]$ and its calculation is independent of the initial conditions. The recursive calculation of the preintegration means that the local increment is updated every time an IMU measurement is acquired. According to equation (\ref{c_b_i}), the local increment $\Upsilon$ can be interpreted as the motion relative to the free-falling frame, which is consistent with the conclusion given in~\cite{fourmy2021contact}.

Equation (\ref{local_increment_recursive}) can be reformulated as 
\begin{equation}\label{IMU_lie_group}
\begin{aligned}
\begin{bmatrix}
\Delta R_j & \Delta v_j & \Delta r_j\\
0_{1\times 3}&1&t_j\\
0_{1\times 3}&0&1
\end{bmatrix}
&=\begin{bmatrix}
\Delta R_i & \Delta v_i & \Delta r_i\\
0_{1\times 3}&1&t_i\\
0_{1\times 3}&0&1
\end{bmatrix}
\begin{bmatrix}
F({i,j}) & \mu({i,j}) & \zeta({i,j})\\
0_{1\times 3}&1&\Delta t_{ij}\\
0_{1\times 3}&0&1
\end{bmatrix}\\
\Upsilon_j'&=\Upsilon_i'\Upsilon_{ij}'
\end{aligned}
\end{equation}
where $t_j=t_i+\Delta t_{ij}$ and the matrix Lie group $\Upsilon'$ is slightly different with the matrix Lie group $\Upsilon$.

Notably, the matrix Lie group $\Upsilon'$ is the same with the compact delta matrix Lie group given in~\cite{fourmy2021contact}. However, the preintegration in~\cite{fourmy2021contact} only considered the non-rotating reference frame, that is, the acceleration are assumed to be constant globally. 
Meanwhile, the rotation of Earth is not taken into account.

In order to obtain the local increment between arbitrary time instance $t_i$ and $t_j$, the recursive formula for $\Upsilon_{ij}$ is given as
\begin{equation}\label{iterative_update_upsilon_ij}
\Upsilon_{ij}=\Phi_{j-1,j}(\Upsilon_{i,j-1})\Upsilon_{j-1,j}
\end{equation}
where the time interval $t_j-t_{j-1}$ in fact denotes the sampling interval of high-rate IMU measurements, which assumed fixed of constant velocity increment and angular increment during the integration interval.
It is worth noting that the iterative calculation of the preintegration measurement is also through the above equation. It is believed that this special recursive formulation makes it easily to extend the preintegration principle to other high-frequency sensors, such as high-frequency contact forces sensors in legged robots~\cite{fourmy2021contact}.
%%%%%%%%%%%%%%%%%%%%%%%%%%%%%%%%%%%%%%%%%%%%%%%%%%%%%%%%%%%
\section{Global Increment on NED frame and ECEF frame}
For the kinematic equations defined on the NED navigation frame~(\ref{differential}) and ECEF frame~(\ref{lie_group_state_differential_eb_e}), similar results can be obtained by introducing an auxiliary velocity transformation.

As the second order kinematic equation on transformed ECEF frame can be obtained by introducing $v_{ib}^e=v_{eb}^e+\omega_{ie}^e\times r_{eb}^e$, we can get the discretization of (\ref{matrix_Lie_group_eb_n}) similar to the discretization of (\ref{new_state_ie_n}) by a reverse process, that is 
\begin{equation}\label{ECEF_composition}
\mathcal{X}_t=\Gamma_t'\Phi_t(\mathcal{X}_0)\Upsilon_t
\end{equation}
where $\Gamma_t'$ is defined as
\begin{equation}\label{Gamma_ecef}
\Gamma_t'=\begin{bmatrix}
\Gamma_t^C & \Gamma_t^v-\omega_{ie}^e\times r_{eb}^e & \Gamma_t^r\\
0_{1\times 3}&1&0\\
0_{1\times 3}&0&1
\end{bmatrix}=\begin{bmatrix} I_{3\times 3} & -\omega_{ie}^e\times r_{eb}^e& 0_{3\times 1}\\
0_{1\times 3} & 0 &0\\0_{1\times 3} & 0 &0
\end{bmatrix}\begin{bmatrix}
\Gamma_t^C & \Gamma_t^v & \Gamma_t^r\\
0_{1\times 3}& 1&0\\
0_{1\times 3}&0&1
\end{bmatrix}
\end{equation}
and quantities in $\Gamma_t$ are defined as same as that in equation (\ref{Gamma_2}). For the velocity quantity, it is easy to verify that 
\begin{equation}\label{new_velocity_ECEF}
\begin{aligned}
&\frac{d}{dt}{\Gamma_t^{'}}^{v}=\frac{d}{dt}\left(\Gamma_t^v-\omega_{ie}^e\times r_{eb}^e\right)=\dot{\Gamma}_t^v-\omega_{ie}^e\times v_{eb}^e=G_{ib}^e-\omega_{ie}^e\times \Gamma_t^v-\omega_{ie}^e\times v_{eb}^e\\
=&g_{ib}^e+\left(\omega_{ie}^e\times\right)^2r_{eb}^e-\omega_{ie}^e\times \Gamma_t^v-\omega_{ie}^e\times v_{eb}^e=-\omega_{ie}^e\times \left(\Gamma_t^v-\omega_{ie}^e\times r_{eb}^e\right)+g_{ib}^e-\omega_{ie}^e\times v_{eb}^e\\
=&-\omega_{ie}^e\times {\Gamma_t^{'}}^{v}+g_{ib}^e-\omega_{ie}^e\times v_{eb}^e
\end{aligned}
\end{equation}

For the position quantity, it is easy to verify that 
\begin{equation}\label{new_position_ECEF}
\frac{d}{dt}{\Gamma_t^r}=\Gamma_t^v={\Gamma_t^{'}}^{v}+\omega_{ie}^e\times r_{eb}^e
\end{equation}

If the term $g_{ib}^e-\omega_{ie}^e\times v_{eb}^e$ and $\omega_{ie}^e\times r_{eb}^e$ can be treated as constant between continuous time interval $[t_i,t_{i+1}]$, then the global increment can be calculated by the approach that has been used in section~\ref{global_increment_section}. The above assumption is reasonable in the practical implementation during two consecutive high frequency measurement timestamps.

Therefore, the navigation state solution to differential equation (\ref{lie_group_state_differential_eb_e}) is given as
\begin{equation}\label{vebe_discrete}
\begin{aligned}
C_{b,t}^e&=\Gamma_t^C C_0 \Delta C\\
v_{eb,t}^e&=\Gamma_t^v+\Gamma_t^C(C_0\Delta v+v_{eb,0}^e)-\omega_{ie}^e\times r_{eb,t}^e\\
r_{eb,t}^e&=\Gamma_t^r+\Gamma_t^C(C_0\Delta r+r_{eb,0}^e+tv_{eb,0}^e)
\end{aligned}
\end{equation}
where $r_{eb,t}^e$ in velocity discretization can be obtained by calculating the position $r_{eb,t}^e$ first.

Similarly, the discretization of (\ref{matrix_Lie_group_eb_n}) can be written as
\begin{equation}\label{NED_composition}
\mathcal{X}_t=\Gamma_t'\Phi_t(\mathcal{X}_0)\Upsilon_t
\end{equation}
where $\Gamma_t'$ is defined as
\begin{equation}\label{Gamma_ned}
\Gamma_t'=\begin{bmatrix}
\Gamma_t^C & \Gamma_t^v-\omega_{ie}^n\times r_{en}^n & \Gamma_t^r\\
0_{1\times 3}&1&0\\
0_{1\times 3}&0&1
\end{bmatrix}=\begin{bmatrix} I_{3\times 3} & -\omega_{ie}^n\times r_{en}^n& 0_{3\times 1}\\
0_{1\times 3} & 0 &0\\0_{1\times 3} & 0 &0
\end{bmatrix}\begin{bmatrix}
\Gamma_t^C & \Gamma_t^v & \Gamma_t^r\\
0_{1\times 3}& 1&0\\
0_{1\times 3}&0&1
\end{bmatrix}
\end{equation}
and quantities in $\Gamma_t$ are defined as same as that in equation (\ref{Gamma_1}). For the velocity quantity, it is easy to verify that 
\begin{equation}\label{new_velocity_NED}
\begin{aligned}
&\frac{d}{dt}{\Gamma_t^{'}}^{v}=\frac{d}{dt}\left(\Gamma_t^v-\omega_{ie}^n\times r_{en}^n\right)
=\dot{\Gamma}_t^v-\frac{d}{dt}(\omega_{ie}^n\times r_{en}^n)=\dot{\Gamma}_t^v-\frac{d}{dt}(C_e^n(\omega_{ie}^e\times) r_{en}^e)\\
=&G_{in}^n-\omega_{in}^n\times \Gamma_t^v+(\omega_{en}^n\times)(\omega_{ie}^n\times)r_{en}^n-\omega_{ie}^n\times v_{en}^n\\
=&g_{in}^n+\left(\omega_{ie}^n\times\right)^2r_{en}^n-\omega_{in}^n\times \Gamma_t^v+(\omega_{en}^n\times)(\omega_{ie}^n\times)r_{en}^n-\omega_{ie}^n\times v_{en}^n\\
=&-\omega_{in}^n\times \left(\Gamma_t^v-\omega_{ie}^n\times r_{en}^n\right)+g_{in}^n-\omega_{ie}^n\times v_{en}^n\\
=&-\omega_{in}^n\times {\Gamma_t^{'}}^{v}+g_{in}^n-\omega_{ie}^n\times v_{en}^n
\end{aligned}
\end{equation}

For the velocity quantity, it is easy to verify that 
\begin{equation}\label{new_position_NED}
\frac{d}{dt}{\Gamma_t^r}=\Gamma_t^v={\Gamma_t^{'}}^{v}+\omega_{ie}^n\times r_{en}^n
\end{equation}

If the term $g_{in}^n-\omega_{ie}^n\times v_{en}^n$ and $\omega_{ie}^n\times r_{en}^n$can be treated as constant between continuous time interval $[t_i,t_{i+1}]$, the differential equations of the global increment in NED navigation frame can also be treated as switched linear system and the calculation procedure in NED navigation frame is similar to that in transformed NED navigation frame.

Therefore, the navigation state solution to differential equation (\ref{differential}) is given as
\begin{equation}\label{venn_discrete}
\begin{aligned}
C_{b,t}^n&=\Gamma_t^C C_0 \Delta C\\
v_{en,t}^n&=\Gamma_t^v+\Gamma_t^C(C_0\Delta v+v_{en,0}^n)-\omega_{ie}^n\times r_{en,t}^n\\
r_{en,t}^n&=\Gamma_t^r+\Gamma_t^C(C_0\Delta r+r_{en,0}^n+tv_{en,0}^n)
\end{aligned}
\end{equation}
where $r_{en,t}^n$ in velocity discretilization can be obtained by calculating the position $r_{en,t}^n$ first.
%%%%%%%%%%%%%%%%%%%%%%%%%%%%%%%%%%%%%%%%%%%%%%%
\section{Batch and Incremental Navigation State Propagation}
Considering the propagation of the navigation state between two arbitrary instants $t_i$ and $t_j$ in the transformed NED navigation frame and transformed ECEF frame, the navigation state can be represented as
\begin{equation}\label{state_propagation}
\begin{aligned}
T_j&\overset{(\ref{decomposition_initial})}{=}\Gamma_j\Phi_j(T_0)\Upsilon_j
\overset{(\ref{global_increment_recursive})}{=}\Gamma_{ij}\Phi_{ij}(\Gamma_i)\Phi_j(T_0)\Upsilon_j
\overset{(\ref{local_increment_recursive})}{=}\Gamma_{ij}\Phi_{ij}(\Gamma_i)\Phi_j(T_0)\Phi_{ij}(\Upsilon_i)\Upsilon_{ij}\\
&\overset{(\ref{automorphism_Phi})}{=}\Gamma_{ij}\Phi_{ij}(\Gamma_i)\Phi_{ij}\left( \Phi_{i}(T_0)\right)\Phi_{ij}(\Upsilon_i)\Upsilon_{ij}
\overset{(\ref{automorphism_Phi})}{=}\Gamma_{ij}\Phi_{ij}\left( \Gamma_i\Phi_{i}(T_0)\Upsilon_i\right)\Upsilon_{ij}
\overset{(\ref{decomposition_initial})}{=}\Gamma_{ij}\Phi_{ij}\left( T_i\right)\Upsilon_{ij}
\end{aligned}
\end{equation}
There is a change from the continuous time to the discrete time in the above equation as we have considered the recursive formulation of the local increment and global increment in discrete time.

Subsequently, the navigation state propagation can be computed iteratively in a closed-form. 
Meanwhile, {$T_j= \Gamma_{ij}\Phi_{ij}(T_i)\Upsilon_{ij}$is used in the recursive calculation of the navigation state,} that is
 \begin{equation}\label{recursive_rotation_velocity_position}
 \begin{aligned}
 C_j&=\Gamma_{ij}^CC_i\Upsilon_{ij}^C\\
 v_j&=\Gamma_{ij}^CC_i\Upsilon_{ij}^v+\Gamma_{ij}^Cv_i+\Gamma_{ij}^v\\
 r_j&=\Gamma_{ij}^CC_i\Upsilon_{ij}^r+\Gamma_{ij}^C(r_i+\Delta t_{ij}v_i)+\Gamma_{ij}^r
 \end{aligned}
 \end{equation}
 
 \begin{remark}
 On the one hand, when the time interval $[t_i,t_j]$ becomes two consecutive measurement timestamps, it can be shown that the above result is equivalent to the mechanization of the strapdown inertial navigation system (SINS) in the transformed ECEF frame and the transformed NED navigation frame. 
 On the other hand, the preintegration measurement is actually contained in the mechanization process of the traditional SINS.
 Therefore, the factor graph based optimization methods for inertial-integrated system improves the accuracy of the SINS by batch smoothing instead of the sequential Bayesian inference in essentially. Compared with the filtering-based method, the improvement is achieved at the expense of increased computation. For inertial-integrated navigation system, if the error compensation algorithms are well chosen, the essential difference between filtering method and factor graph method is only that the batch estimation is adopted by factor graph, while the recursive estimation is adopted by filtering.
  \end{remark}

According to equation (\ref{recursive_rotation_velocity_position}), the recursive calculation of the navigation state between two consecutive time instance $[t_i,t_{i+1}]$ is given as 
 \begin{equation}\label{recursive_rotation_velocity_position_consecutive}
\begin{aligned}
C_{i+1}&=\Gamma_{i,i+1}^CC_i\Upsilon_{i,i+1}^C\\
v_{i+1}&=\Gamma_{i,i+1}^CC_i\Upsilon_{i,i+1}^v+\Gamma_{i,i+1}^Cv_i+\Gamma_{i,i+1}^v\\
r_{i+1}&=\Gamma_{i,i+1}^CC_i\Upsilon_{i,i+1}^r+\Gamma_{i,i+1}^C(r_i+\Delta t_{i,i+1}v_i)+\Gamma_{i,i+1}^r
\end{aligned}
\end{equation}

\begin{remark}
	It is worth noting that the above discretization of the navigation state in consecutive time is equivalent to the mechanization process of the traditional SINS and it can be shown by some simple mathematical manipulations. Therefore, the preintegration theory of IMU in different frames can be used for the factor graph-based framework without losing accuracy. Of course, it does not improve the accuracy of the system when compared with filtering based framework from the IMU's perspective.
\end{remark}

According to the velocity update and position update in equation (\ref{recursive_rotation_velocity_position}), the initial attitude can be calculated as
\begin{equation}\label{OBA_initial_alignment}
\begin{aligned}
C_i\Upsilon_{ij}^v&={\Gamma_{ij}^C}^T\left(v_j-\Gamma_{ij}^Cv_i-\Gamma_{ij}^v\right)\\
C_i\Upsilon_{ij}^r&={\Gamma_{ij}^C}^T\left(r_{j}-\Gamma_{ij}^C(r_i+\Delta t_{ij}v_i)-\Gamma_{ij}^r\right)
\end{aligned}
\end{equation}
It is worth noting that the above form is the same with the problem of optimization based coarse alignment method~\cite{wu2013velocity, chang2016optimization}.

Once the global increment is calculated, the local increment between two timestamps $t_i$ and $t_j$ can be given as
\begin{equation}\label{local_increment_gamma}
\Upsilon_{ij}=(\Gamma_{ij}\Phi_{ij}(T_i))^{-1}T_j
\end{equation}
It is preintegration factors that relates to the navigation states. 

Finally, we can obtain 
\begin{equation}\label{quantities_rotation_velocity_position}
\begin{aligned}
\Upsilon_{ij}^C&=(\Gamma_{ij}^CC_i)^{-1}C_j\\
\Upsilon_{ij}^v&=C_i^T\left({\Gamma_{ij}^C}^T(v_j-\Gamma_{ij}^v )-v_i)\right)\\
\Upsilon_{ij}^r&=C_i^T\left({\Gamma_{ij}^C}^T(r_j-\Gamma_{ij}^r )-(r_i+\Delta t_{ij}v_i)\right)
\end{aligned}
\end{equation}

Meanwhile, the uncertainty of the local increment between the consecutive sensor time steps $[t_i,t_{i+1}]$ can be represented as
\begin{equation}\label{uncertainty_Lie_group_ii1}
\Upsilon_{i,i+1}=\hat{\Upsilon}_{i,i+1}Exp(\eta_{i,i+1}),\eta_{i,i+1}=G_i\nu_i\sim \mathcal{N}(0_{9\times 1},G_{i}cov(\nu_{i})G_{i}^T),\nu_{i}=\begin{bmatrix}
\eta^{\omega} \\ \eta^{a}
\end{bmatrix}
\end{equation}

For the preintegration calculated by assuming piecewise IMU measurements, the preintegration measurement can be represented as
\begin{equation}\label{Estimation_accurate1}
\Upsilon_{i,i+1}=\begin{bmatrix}
F(i,i+1) & \mu_1({i,i+1}) & \zeta_1({i,i+1})\\
0_{1\times 3} & 1 &0\\
0_{1\times 3} & 0 &1
\end{bmatrix}=\begin{bmatrix}
Exp(\omega_{ib}^b\Delta t) & f_{ib}^b\Delta t & \frac{1}{2}f_{ib}^b\Delta
 t^2\\
 0_{1\times 3} & 1 &0\\
 0_{1\times 3} & 0 &1
\end{bmatrix}
\end{equation}

For the preintegration calculated by assuming piecewise constant global acceleration, the preintegration measurement can be represented as
\begin{equation}\label{Estimation_accurate2}
\begin{aligned}
\Upsilon_{i,i+1}=&\begin{bmatrix}
F(i,i+1) & \mu_2({i,i+1}) & \zeta_2({i,i+1})\\
0_{1\times 3} & 1 &0\\
0_{1\times 3} & 0 &1
\end{bmatrix}\\
=&\begin{bmatrix}
\Gamma_{0}(\omega_{ib}^b\Delta t) & \Gamma_{1}(\omega_{ib}^b\Delta t)f_{ib}^b\Delta t & \Gamma_{2}(\omega_{ib}^b\Delta t)f_{ib}^b\Delta
t^2\\
0_{1\times 3} & 1 &0\\
0_{1\times 3} & 0 &1
\end{bmatrix}
\end{aligned}
\end{equation}

For the preintegration calculated when the multi-sample error compensation is taking into consideration, the preintegration measurement can be represented as
\begin{equation}\label{Estimation_accurate3}
\begin{aligned}
\Upsilon_{i,i+1}=&\begin{bmatrix}
\mathcal{F}(i,i+1) & \mu_3({i,i+1}) & \zeta_3({i,i+1})\\
0_{1\times 3} & 1 &0\\
0_{1\times 3} & 0 &1
\end{bmatrix}
=\begin{bmatrix}
\Gamma_{0}(\omega_{ib}^b\Delta t) & \delta v_{(i)} & 0.5 \delta v _{(i)}\Delta t\\
0_{1\times 3} & 1 &0\\
0_{1\times 3} & 0 &1
\end{bmatrix}
\end{aligned}
\end{equation}

We denote the estimated of the preintegration $\Upsilon_{i,i+1}$ as 
\begin{equation}\label{Estimation_accurate2_estimeated}
\begin{aligned}
\hat{\Upsilon}_{i,i+1}=&\begin{bmatrix}
\Gamma_{0}((\hat{\omega}_{ib}^b-b_i^w)\Delta t) & \Gamma_{1}((\hat{\omega}_{ib}^b-b_i^w)\Delta t)(\hat{f}_{ib}^b-b_i^a)\Delta t & \Gamma_{2}((\hat{\omega}_{ib}^b-b_i^w)\Delta t)(\hat{f}_{ib}^b-b_i^a)\Delta
t^2\\
0_{1\times 3} & 1 &0\\
0_{1\times 3} & 0 &1
\end{bmatrix}\\
=&\begin{bmatrix}
\hat{\Gamma}_{0} & \hat{\Gamma}_{1}\cdot (\hat{f}_{ib}^b-b_i^a)\Delta t  & \hat{\Gamma}_{2}\cdot(\hat{f}_{ib}^b-b_i^a)\Delta
t^2\\
0_{1\times 3} & 1 &0\\
0_{1\times 3} & 0 &1
\end{bmatrix}
\end{aligned}
\end{equation}

In order to compute the Jacobian matrix $G_i$ defined in equation (\ref{uncertainty_Lie_group_ii1}), we assume that 
\begin{equation}\label{uncertainty_Lie_group_ii1_G_i}
\Upsilon_{i,i+1}=\hat{\Upsilon}_{i,i+1}Exp\left( \begin{bmatrix}
A\\B\\C
\end{bmatrix}\right)=\hat{\Upsilon}_{i,i+1}\begin{bmatrix}
\Gamma_{0}(A) & \Gamma_{1}(A)B &  \Gamma_{1}(A)C\\
0_{1\times 3} & 1 &0\\
0_{1\times 3} & 0 &1
\end{bmatrix}
\end{equation}
If the biases are assumed to known during the time interval $[t_i,t_j]$ and assumed to be $(b_i^w, b_i^a)$, then $A$ can be computed by the first order Taylor series as
\begin{equation}\label{A}
\begin{aligned}
&A=Log(\Gamma_{0}((\hat{\omega}_{ib}^b-b_i^w)\Delta t)^{-1}\Gamma_{0}(\omega_{ib}^b\Delta t))\\
\overset{(\ref{BCH_1})}{=}&Log(\Gamma_{0}((\hat{\omega}_{ib}^b-b_i^w)\Delta t)^{-1}\Gamma_{0}((\hat{\omega}_{ib}^b-b_i^w)\Delta t))\Gamma_{0}(\Gamma_{1}(-(\hat{\omega}_{ib}^b-b_i^w)\Delta t)(-\eta^{\omega}\Delta t)))\\
=&\Gamma_{1}(-(\hat{\omega}_{ib}^b-b_i^w)\Delta t)(-\eta^{\omega}\Delta t)=-\Gamma_{1}(-(\hat{\omega}_{ib}^b-b_i^w)\Delta t)\Delta t \eta^{\omega}\\
=&-\hat{\Gamma}_{1}^-\cdot \Delta t \eta^{\omega}=-\hat{\Gamma}_{0}^{-1}\cdot\hat{\Gamma}_{0}\cdot\hat{\Gamma}_{1}^-\cdot \Delta t=-\hat{\Gamma}_{0}^{-1}\cdot\hat{\Gamma}_{1}\cdot \Delta t \eta^{\omega}
\end{aligned}
\end{equation}
where $\Gamma_{1}(-(\hat{\omega}_{ib}^b-b_i^w)\Delta t)$ is denoted as $\hat{\Gamma}_{1}^-$, the verification of $\hat{\Gamma}_{0}\cdot\hat{\Gamma}_{1}^-=\hat{\Gamma}_{1}$ can be found in \cite{luo2020geometry}.

As $A$ is a small quantity, $\Gamma_{1}(A)$ can be approximated as $\Gamma_{1}(A)\approx I$. Therefore, $B$ can be computed as 
\begin{equation}\label{B}
\begin{aligned}
&B=\Gamma_{0}((\hat{\omega}_{ib}^b-b_i^w)\Delta t)^{-1}\left(\Gamma_{1}(\omega_{ib}^b\Delta t)f_{ib}^b\Delta t -\Gamma_{1}((\hat{\omega}_{ib}^b-b_i^w)\Delta t)(\hat{f}_{ib}^b-b_i^a)\Delta t    \right)\\
\overset{(\ref{BCH_1})}{=}&\hat{\Gamma}_{0}^{-1}\cdot \left(\Gamma_{1}((\hat{\omega}_{ib}^b-b_i^w)\Delta t)\Gamma_{0}\left(\Gamma_{2}(-(\hat{\omega}_{ib}^b-b_i^w)\Delta t) (-\eta^{\omega}\Delta t)\right)  f_{ib}^b\Delta t -\hat{\Gamma}_{1}\cdot (\hat{f}_{ib}^b-b_i^a)\Delta t    \right)\\
\approx &\hat{\Gamma}_{0}^{-1}\cdot \left( \hat{\Gamma}_{1}\cdot f_{ib}^b\Delta t-  \hat{\Gamma}_{1}\cdot (\hat{f}_{ib}^b-b_i^a) \Delta t +\hat{\Gamma}_{1}\cdot \left(\hat{\Gamma}_{2}^-\cdot (-\eta^{\omega}\Delta t)  \right)\times  f_{ib}^b\Delta t  \right)\\
=&\hat{\Gamma}_{0}^{-1}\cdot \left(\hat{\Gamma}_{1}\cdot  (-\eta^a\Delta t)+\hat{\Gamma}_{1}\cdot(f_{ib}^b\times) \hat{\Gamma}_{2}^-\cdot (\eta^{\omega}\Delta t)\Delta t\right)\\
=&\hat{\Gamma}_{0}^{-1}\cdot \left(-\hat{\Gamma}_{1}\cdot \Delta t \eta^a+\hat{\Gamma}_{1}\cdot ((\hat{f}_{ib}^b-b_i^a)\times ) \hat{\Gamma}_{2}^-\cdot\Delta t^2 \eta^{\omega}\right)
\end{aligned}
\end{equation}
where $\Gamma_{2}(-(\hat{\omega}_{ib}^b-b_i^w)\Delta t)$ is denoted as $\hat{\Gamma}_{2}^-$.

Similarly, $C$ can be computed as 
\begin{equation}\label{C}
\begin{aligned}
&C=\Gamma_{0}((\hat{\omega}_{ib}^b-b_i^w)\Delta t)^{-1}\left(\Gamma_{2}({\omega}_{ib}^b \Delta t)f_{ib}^b\Delta t^2 -\Gamma_{2}((\hat{\omega}_{ib}^b-b_i^w) \Delta t)(\hat{f}_{ib}^b-b_i^a)\Delta t^2    \right)\\
\overset{(\ref{BCH_1})}{=}&\hat{\Gamma}_{0}^{-1}\cdot \left(\Gamma_{2}((\hat{\omega}_{ib}^b-b_i^w)\Delta t)\Gamma_{0}\left(\Gamma_{3}(-(\hat{\omega}_{ib}^b-b_i^w)\Delta t) (-\eta^{\omega}\Delta t)\right)  f_{ib}^b\Delta t^2 -\hat{\Gamma}_{2}\cdot (\hat{f}_{ib}^b-b_i^a)\Delta t^2    \right)\\
\approx &\hat{\Gamma}_{0}^{-1}\cdot \left( \hat{\Gamma}_{2}\cdot f_{ib}^b\Delta t^2-  \hat{\Gamma}_{2}\cdot (\hat{f}_{ib}^b-b_i^a) \Delta t^2 +\hat{\Gamma}_{2}\cdot \left(\hat{\Gamma}_{3}^-\cdot (-\eta^{\omega}\Delta t)  \right)\times  f_{ib}^b\Delta t^2  \right)\\
=&\hat{\Gamma}_{0}^{-1}\cdot \left(\hat{\Gamma}_{2}\cdot  (-\eta^a\Delta t^2)+\hat{\Gamma}_{2}\cdot(f_{ib}^b\times) \hat{\Gamma}_{3}^-\cdot (\eta^{\omega}\Delta t)\Delta t^2\right)\\
=&\hat{\Gamma}_{0}^{-1}\cdot \left(-\hat{\Gamma}_{2}\cdot \Delta t^2 \eta^a+\hat{\Gamma}_{2}\cdot ((\hat{f}_{ib}^b-b_i^a)\times ) \hat{\Gamma}_{3}^-\cdot\Delta t^3 \eta^{\omega}\right)
\end{aligned}
\end{equation}
where $\Gamma_{3}(-(\hat{\omega}_{ib}^b-b_i^w)\Delta t)$ is denoted as $\hat{\Gamma}_{3}^-$.

We thus obtain the Jacobian matrix $G_i$ with respect to noise as
\begin{equation}\label{G_i_1}
G_i=\begin{bmatrix}
-\hat{\Gamma}_{0}^{-1}\cdot\hat{\Gamma}_{1}\cdot \Delta t  & 0_{3\times 3}\\
\hat{\Gamma}_{0}^{-1}\cdot  \hat{\Gamma}_{1}\cdot ((\hat{f}_{ib}^b-b_i^a)\times ) \hat{\Gamma}_{2}^-\cdot \Delta t^2&-\hat{\Gamma}_{0}^{-1}\cdot\hat{\Gamma}_{1}\cdot\Delta t\\
\hat{\Gamma}_{0}^{-1}\cdot \hat{\Gamma}_{2}\cdot((\hat{f}_{ib}^b-b_i^a)\times ) \hat{\Gamma}_{3}^-\cdot\Delta t^3& -\hat{\Gamma}_{0}^{-1}\cdot\hat{\Gamma}_{2}\cdot\Delta t^2
\end{bmatrix}
\end{equation}

If the central Gaussian distribution of the navigation state $T_i$ is defined based right perturbation:
\begin{equation}\label{CGdistribution}
T_i:=\hat{T}_i Exp(\xi_i),\xi_i\sim N(0,\Sigma_i)
\end{equation}
The propagation of $T_i$ through noisy model between two consecutive time steps can be obtained as
\begin{equation}\label{propagation_noisy_model}
\begin{aligned}
T_{i+1}&=\Gamma_{i,i+1}\Phi_{i,i+1}(T_i)\Upsilon_{i,i+1}=\Gamma_{i,i+1}\Phi_{i,i+1}(\hat{T}_iExp(\xi_i))\hat{\Upsilon}_{i,i+1}Exp(\eta_{i,i+1})\\
&=\Gamma_{i,i+1}\Phi_{i,i+1}(\hat{T}_i)\hat{\Upsilon}_{i,i+1}\hat{\Upsilon}_{i,i+1}^{-1}\Phi_{i,i+1}(Exp(\xi_i))\hat{\Upsilon}_{i,i+1}Exp(\eta_{i,i+1})\\
&=\hat{T}_{i+1} Ad_{\hat{\Upsilon}_{i,i+1}^{-1}}\left(Exp(F\xi_i) \right)Exp(\eta_{i,i+1})=\hat{T}_{i+1}Exp(Ad_{\hat{\Upsilon}_{i,i+1}^{-1}}F\xi_i)Exp(\eta_{i,i+1})\\
&\overset{(\ref{CGdistribution})}{=}\hat{T}_{i+1}Exp(\xi_{i+1})
\end{aligned}
\end{equation}
where $\xi_{i+1}$ is assumed to satisfy $\xi_{i+1}\sim N(0,\Sigma_{i+1})$. As $\xi_i$ and $\eta_{i,i+1}$ all are small quantities, the BCH formula of the matrix Lie group is applied and we can get 
\begin{equation}\label{BCH_approximation}
\xi_{i+1}=Log\left(Exp(Ad_{\hat{\Upsilon}_{i,i+1}^{-1}}F\xi_i)Exp(\eta_{i,i+1}) \right)\approx \underbrace{Ad_{\hat{\Upsilon}_{i,i+1}^{-1}}F}_{A_i}\xi_i+\eta_{i,i+1}\triangleq A_i\xi_i+\eta_{i,i+1}
\end{equation}
where $A_i$ can be calculated as
\begin{equation}\label{A_i}
\begin{aligned}
&A_i=Ad_{\hat{\Upsilon}_{i,i+1}^{-1}}F=Ad_{\hat{\Upsilon}_{i,i+1}}^{-1}F\\
\overset{(\ref{Estimation_accurate2})}{=}&\begin{bmatrix}
\hat{\Gamma}_{0}^{-1} &  0_{3\times 3}& 0_{3\times 3}\\
-\hat{\Gamma}_{0}^{-1}\cdot \left(\hat{\Gamma}_{1}\cdot (\hat{f}_{ib}^b-b_i^a)\Delta t\right)\times &\hat{\Gamma}_{0}^{-1}& 0_{3\times 3}\\ -\hat{\Gamma}_{0}^{-1}\cdot\left(\hat{\Gamma}_{2}\cdot (\hat{f}_{ib}^b-b_i^a)\Delta
t^2\right)\times & 0_{3\times 3}&\hat{\Gamma}_{0}^{-1}
\end{bmatrix}F\\
\overset{(\ref{log_linearity_property})}{=}&\begin{bmatrix}
\hat{\Gamma}_{0}^{-1} &  0_{3\times 3}& 0_{3\times 3}\\
-\hat{\Gamma}_{0}^{-1}\cdot \left(\hat{\Gamma}_{1}\cdot (\hat{f}_{ib}^b-b_i^a)\Delta t\right)\times &\hat{\Gamma}_{0}^{-1}& 0_{3\times 3}\\ -\hat{\Gamma}_{0}^{-1}\cdot\left(\hat{\Gamma}_{2}\cdot (\hat{f}_{ib}^b-b_i^a)\Delta
t^2\right)\times & 0_{3\times 3}&\hat{\Gamma}_{0}^{-1}
\end{bmatrix}\begin{bmatrix}
I_{3\times 3} &  0_{3\times 3}&  0_{3\times 3}\\
 0_{3\times 3}& I_{3\times 3} &  0_{3\times 3}\\
 0_{3\times 3}& \Delta tI& I_{3\times 3}
\end{bmatrix}\\
=&\begin{bmatrix}
\hat{\Gamma}_{0}^{-1} &  0_{3\times 3}& 0_{3\times 3}\\
-\hat{\Gamma}_{0}^{-1}\cdot \left(\hat{\Gamma}_{1}\cdot (\hat{f}_{ib}^b-b_i^a)\Delta t\right)\times &\hat{\Gamma}_{0}^{-1}& 0_{3\times 3}\\ -\hat{\Gamma}_{0}^{-1}\cdot\left(\hat{\Gamma}_{2}\cdot (\hat{f}_{ib}^b-b_i^a)\Delta
t^2\right)\times &\Delta t  \hat{\Gamma}_{0}^{-1}&\hat{\Gamma}_{0}^{-1}
\end{bmatrix}\\
=&\hat{\Gamma}_{0}^{-1}\begin{bmatrix}
I_{3\times 3} &  0_{3\times 3}& 0_{3\times 3}\\
- \left(\hat{\Gamma}_{1}\cdot (\hat{f}_{ib}^b-b_i^a)\Delta t\right)\times &I_{3\times 3}& 0_{3\times 3}\\ -\left(\hat{\Gamma}_{2}\cdot (\hat{f}_{ib}^b-b_i^a)\Delta
t^2\right)\times &\Delta t  I_{3\times 3}&I_{3\times 3}
\end{bmatrix}
\end{aligned}
\end{equation}

Equation (\ref{A_i}) and equation (\ref{G_i_1}) provide the close-form solutions for $A_i$ and $G_i$ in a discrete manner and will lead to a more accurate uncertainty propagation.
Therefore, the covariance of the discrepancy evolves as
\begin{equation}\label{covariance_propogation}
\Sigma_{i+1}=A_i\Sigma_iA_i^T+G_iQ_iG_i^T
\end{equation}
where $Q_i=cov(\nu_i)$ is the input noise covariance and can be treated as constant matrix $Q$. 
The iterative updating of the covariance matrix will provide convenient computation for online estimation when a new IMU measurement is obtained.

However, it is obvious that the matrix $A_i$ is dependent on the specific force which may be noisy in low-cost IMU. Therefore, the common frame error representation can be used to cancel this term\cite{scherzinger1994modified}. As the right invariant error will lead the common frame error representation, which means the central Gaussian distribution of the navigation state $T_i$ is supposed to be defined based on left perturbation:
\begin{equation}\label{CGdistribution_left}
T_i:= Exp(\xi_i)\hat{T}_i,\xi_i\sim N(0,\Sigma_i)
\end{equation}
The propagation of $T_i$ through noisy model between two consecutive time steps can be obtained as
\begin{equation}\label{propagation_noisy_model_left}
\begin{aligned}
T_{i+1}&=\Gamma_{i,i+1}\Phi_{i,i+1}(T_i)\Upsilon_{i,i+1}=\Gamma_{i,i+1}\Phi_{i,i+1}(Exp(\xi_i)\hat{T}_i)\hat{\Upsilon}_{i,i+1}Exp(\eta_{i,i+1})\\
&=\Gamma_{i,i+1}\Phi_{i,i+1}(Exp(\xi_i))\Gamma_{i,i+1}^{-1}\Gamma_{i,i+1}\Phi_{i,i+1}(\hat{T}_i)\hat{\Upsilon}_{i,i+1}Exp(\eta_{i,i+1})\\
&= Ad_{\Gamma_{i,i+1}}\left(Exp(F\xi_i) \right)\hat{T}_{i+1}Exp(\eta_{i,i+1})\hat{T}_{i+1}^{-1}\hat{T}_{i+1}\\
&=Exp(Ad_{\Gamma_{i,i+1}}F\xi_i)Exp(Ad_{\hat{T}_{i+1}}\eta_{i,i+1})\hat{T}_{i+1}\\
&\overset{(\ref{CGdistribution_left})}{=}Exp(\xi_{i+1})\hat{T}_{i+1}
\end{aligned}
\end{equation}
where $\xi_{i+1}$ is assumed to satisfy $\xi_{i+1}\sim N(0,\Sigma_{i+1})$. As $\xi_i$ and $\eta_{i,i+1}$ all are small quantities, the BCH formula of the matrix Lie group is applied and we can get 
\begin{equation}\label{BCH_approximation_left}
\begin{aligned}
\xi_{i+1}=&Log\left(Exp(Ad_{\Gamma_{i,i+1}}F\xi_i)Exp(Ad_{\hat{T}_{i+1}}\eta_{i,i+1})) \right)\\
\approx &\underbrace{Ad_{\Gamma_{i,i+1}}F}_{A_i}\xi_i+Ad_{\hat{T}_{i+1}}\eta_{i,i+1}\triangleq A_i\xi_i+Ad_{\hat{T}_{i+1}}\eta_{i,i+1}
\end{aligned}
\end{equation}
where $A_i$ can be calculated as
\begin{equation}\label{A_i_left}
\begin{aligned}
A_i=&Ad_{\Gamma_{i,i+1}}F
\overset{(\ref{Gamma_ij_C_v_r})}{=}\begin{bmatrix}
\Gamma_{i,i+1}^C &  0_{3\times 3}& 0_{3\times 3}\\
 \left(\Gamma_{i,i+1}^v\right)\times \Gamma_{i,i+1}^C &\Gamma_{i,i+1}^C& 0_{3\times 3}\\ \left(\Gamma_{i,i+1}^r\right) \times \Gamma_{i,i+1}^C  & 0_{3\times 3}&\Gamma_{i,i+1}^C
\end{bmatrix}\begin{bmatrix}
I_{3\times 3} &  0_{3\times 3}&  0_{3\times 3}\\
 0_{3\times 3}& I_{3\times 3} &  0_{3\times 3}\\
 0_{3\times 3}& \Delta tI_{3\times 3}& I_{3\times 3}
\end{bmatrix}\\
=&\begin{bmatrix}
\Gamma_{i,i+1}^C &  0_{3\times 3}& 0_{3\times 3}\\
\left(\Gamma_{i,i+1}^v\right)\times \Gamma_{i,i+1}^C &\Gamma_{i,i+1}^C& 0_{3\times 3}\\ \left(\Gamma_{i,i+1}^r\right) \times \Gamma_{i,i+1}^C  &\Delta t\Gamma_{i,i+1}^C &\Gamma_{i,i+1}^C
\end{bmatrix}
\end{aligned}
\end{equation}
It is obvious that the matrix $A_i$ is independent of the specific force due to the left perturbation.

Therefore, the covariance of the discrepancy evolves as
\begin{equation}\label{covariance_propogation_left}
\Sigma_{i+1}=A_i\Sigma_iA_i^T+Ad_{\hat{T}_{i+1}}G_iQ_iG_i^TAd_{\hat{T}_{i+1}}^T
\end{equation}

For the transformed ECEF frame, the matrix $A_i$ and $Ad_{\hat{T}_{i+1}}G_i$ are given as
\begin{equation}\label{A_i_transformed_ECEF}
A_i=\begin{bmatrix}
Exp(-\omega_{ie}^e \Delta t_{ij})&  0_{3\times 3}& 0_{3\times 3}\\
\left(\Gamma_{1}(-\omega_{ie}^e\Delta t_{ij})\Delta t_{ij} G_{ib}^e\right)\times Exp(-\omega_{ie}^e \Delta t_{ij}) &Exp(-\omega_{ie}^e \Delta t_{ij})& 0_{3\times 3}\\ 
\left(\Gamma_{2}(-\omega_{ie}^e\Delta t_{ij})\Delta t_{ij}^2 G_{ib}^e\right) \times Exp(-\omega_{ie}^e \Delta t_{ij})  &\Delta tExp(-\omega_{ie}^e \Delta t_{ij}) &Exp(-\omega_{ie}^e \Delta t_{ij})
\end{bmatrix}
\end{equation}
\begin{equation}\label{G_i_right_invariant}
\begin{aligned}
&Ad_{\hat{T}_{i+1}}G_i\overset{(\ref{G_i_1})}{=}
\begin{bmatrix}
\hat{C}_{i+1} & 0_{3\times 3} & 0_{3\times 3}\\
(\hat{v}_{i+1}\times )\hat{C}_{i+1} & \hat{C}_{i+1} & 0_{3\times 3}\\
(\hat{r}_{i+1}\times )\hat{C}_{i+1} & 0_{3\times 3} & \hat{C}_{i+1}
\end{bmatrix}
\hat{\Gamma}_{0}^{-1}\cdot\begin{bmatrix}
-\hat{\Gamma}_{1}\cdot \Delta t  & 0_{3\times 3}\\
  \hat{\Gamma}_{1}\cdot ((\hat{f}_{ib}^b-b_i^a)\times ) \hat{\Gamma}_{2}^-\cdot \Delta t^2&-\hat{\Gamma}_{1}\cdot\Delta t\\
 \hat{\Gamma}_{2}\cdot((\hat{f}_{ib}^b-b_i^a)\times ) \hat{\Gamma}_{3}^-\cdot\Delta t^3& -\hat{\Gamma}_{2}\cdot\Delta t^2
\end{bmatrix}\\
=&\begin{bmatrix}
\hat{C}_{i+1}\hat{\Gamma}_{0}^{-1} & 0_{3\times 3} & 0_{3\times 3}\\
(\hat{v}_{i+1}\times )\hat{C}_{i+1}\hat{\Gamma}_{0}^{-1} & \hat{C}_{i+1}\hat{\Gamma}_{0}^{-1} & 0_{3\times 3}\\
(\hat{r}_{i+1}\times )\hat{C}_{i+1}\hat{\Gamma}_{0}^{-1} & 0_{3\times 3} & \hat{C}_{i+1}\hat{\Gamma}_{0}^{-1}
\end{bmatrix}
\begin{bmatrix}
-\hat{\Gamma}_{1}\cdot \Delta t  & 0_{3\times 3}\\
\hat{\Gamma}_{1}\cdot ((\hat{f}_{ib}^b-b_i^a)\times ) \hat{\Gamma}_{2}^-\cdot \Delta t^2&-\hat{\Gamma}_{1}\cdot\Delta t\\
\hat{\Gamma}_{2}\cdot((\hat{f}_{ib}^b-b_i^a)\times ) \hat{\Gamma}_{3}^-\cdot\Delta t^3& -\hat{\Gamma}_{2}\cdot\Delta t^2
\end{bmatrix}\\
\overset{(\ref{recursive_rotation_velocity_position_consecutive})}{=}&\begin{bmatrix}
\Gamma_{i,i+1}^C\hat{C}_{i} & 0_{3\times 3} & 0_{3\times 3}\\
(\hat{v}_{i+1}\times )\Gamma_{i,i+1}^C\hat{C}_{i} & \Gamma_{i,i+1}^C\hat{C}_{i} & 0_{3\times 3}\\
(\hat{r}_{i+1}\times )\Gamma_{i,i+1}^C\hat{C}_{i} & 0_{3\times 3} &\Gamma_{i,i+1}^C \hat{C}_{i}
\end{bmatrix}
\begin{bmatrix}
-\hat{\Gamma}_{1}\cdot \Delta t  & 0_{3\times 3}\\
\hat{\Gamma}_{1}\cdot ((\hat{f}_{ib}^b-b_i^a)\times ) \hat{\Gamma}_{2}^-\cdot \Delta t^2&-\hat{\Gamma}_{1}\cdot\Delta t\\
\hat{\Gamma}_{2}\cdot((\hat{f}_{ib}^b-b_i^a)\times ) \hat{\Gamma}_{3}^-\cdot\Delta t^3& -\hat{\Gamma}_{2}\cdot\Delta t^2
\end{bmatrix}\\
=&\begin{bmatrix}
-\Gamma_{i,i+1}^C\hat{C}_{i}\hat{\Gamma}_{1}\cdot \Delta t  & 0_{3\times 3}\\
-(\hat{v}_{i+1}\times )\Gamma_{i,i+1}^C\hat{C}_{i}\hat{\Gamma}_{1}\cdot \Delta t+\Gamma_{i,i+1}^C\hat{C}_{i}\hat{\Gamma}_{1}\cdot ((\hat{f}_{ib}^b-b_i^a)\times ) \hat{\Gamma}_{2}^-\cdot \Delta t^2&-\Gamma_{i,i+1}^C\hat{C}_{i}\hat{\Gamma}_{1}\cdot\Delta t\\
-(\hat{r}_{i+1}\times )\Gamma_{i,i+1}^C\hat{C}_{i}\hat{\Gamma}_{1}\cdot \Delta t+\Gamma_{i,i+1}^C\hat{C}_{i}\hat{\Gamma}_{2}\cdot((\hat{f}_{ib}^b-b_i^a)\times ) \hat{\Gamma}_{3}^-\cdot\Delta t^3& -\Gamma_{i,i+1}^C\hat{C}_{i}\hat{\Gamma}_{2}\cdot\Delta t^2
\end{bmatrix}
\end{aligned}
\end{equation}
\begin{remark}
	It is worth noting that the equation (\ref{A_i}) and (\ref{A_i_left}) can be treated as the closed-form error-state transition matrix for the left-invariant EKF and right invariant EKF~\cite{barrau2015non}. There is no approximation about the attitude error although the BCH formula has been used many times. The small quantities used in BCH is due to the small time interval. Therefore, the covariance propagation in equation (\ref{covariance_propogation}) and equation (\ref{covariance_propogation_left}) is more straightforward and accurate than the traditional method.
\end{remark}

For the recursive formula between time instances $t_i$ and $t_j$, the uncertainty $\xi_{j}$ can be obtained recursively
\begin{equation}\label{uncertainty_j}
\begin{aligned}
\xi_{j}&=A_{j-1}\xi_{j-1}+\eta_{j-1,j}\\
&=A_{j-1}A_{j-2}\xi_{j-2}+A_{j-1}\eta_{j-2,j-1}+\eta_{j-1,j}=\cdots\\
&=A_{j-1}A_{j-2}\cdots A_i\xi_i+A_{j-1}A_{j-2}\cdots A_{i+1}\eta_{i,i+1}+A_{j-1}A_{j-2}\cdots A_{i+2}\eta_{i+1,i+2}\\
&\qquad+\cdots +A_{j-1}\eta_{j-2,j-1}+\eta_{j-1,j} \\
&=A_{i}^{j-1}\xi_i+\sum_{k=i}^{j-1}A_{k+1}^{j-1}\eta_{k,k+1}
\end{aligned}
\end{equation}
where $A_i^j$ is defined as
\begin{equation}\label{A_i_j}
A_i^j=\begin{cases}
A_jA_{j-1}\cdots A_i, & j>i \\
A_i, & j=i \\
I,& j<i \\
\end{cases}
\end{equation}

Based on equation (\ref{uncertainty_j}), the recursion of the covariance is given as
\begin{equation}\label{covariance_recursive}
\Sigma_j=A_{i}^{j-1}\Sigma_i {A_{i}^{j-1}}^T+\sum_{k=i}^{j-1}A_{k+1}^{j-1}G_kQG_k^T {A_{k+1}^{j-1}}^T
\end{equation}

Based on the uncertainty representation on the matrix Lie group, the uncertainty of the preintegration measurement is encoded by matrix Lie group exponential mapping, that is
\begin{equation}\label{uncertainty_Lie_group}
\Upsilon_{ij}=\hat{\Upsilon}_{ij}Exp(\eta_{ij})
\end{equation}

Meanwhile, the uncertainty of the local increment $\Upsilon_i$ is assumed to be zero mean central Gaussian distribution, that is
\begin{equation}\label{local_increment_uncertainty}
\begin{aligned}
\Upsilon_j&=\Phi_{j-1,j}(\Upsilon_{j-1})\Upsilon_{j-1,j}=\Phi_{j-1,j}(\hat{\Upsilon}_{j-1}Exp(\eta_{j-1}))\hat{\Upsilon}_{j-1,j}Exp(\eta_{j-1,j})\\
&=\Phi_{j-1,j}(\hat{\Upsilon}_{j-1})\hat{\Upsilon}_{j-1,j}Exp(Ad_{\hat{\Upsilon}_{j-1,j}}F\eta_{j-1})Exp(\eta_{j-1,j})
=\hat{\Upsilon}_jExp(\eta_j)
\end{aligned}
\end{equation}

Therefore, the recursive formula of the noise is given as
\begin{equation}\label{local_increment_uncertainty_recursive}
\begin{aligned}
\eta_j&=Ad_{\hat{\Upsilon}_{j-1,j}}F\eta_{j-1}+\eta_{j-1,j}=A_{j-1}\eta_{j-1}+\eta_{j-1,j}\\
&=A_{j-1}\left( A_{j-2}\eta_{j-2}+\eta_{j-2,j-1}      \right)+\eta_{j-1,j}
=A_i^{j-1}\eta_{i}+\sum_{k=i}^{j-1}A_{k+1}^{j-1}\eta_{k,k+1}
\end{aligned}
\end{equation}
which is same with equation (\ref{uncertainty_j}). It is obvious that the noise propagation of the integration measurement is same with that of the noisy navigation state up to second-order accuracy. 
\section{Analytic Bias Update}
As we have assumed that the biases are exact and fixed during the interval $[t_i,t_j]$, the prejntegration measurements will be recalculated once the biases are changed after the optimization, which is computational expensive.
In order to solve this problem, a first-order Taylor approximate update method of the preintegration measurement with biases change is proposed by linearization in the Lie exponential coordinates.
Assume $\overline{\Upsilon}_{ij}(\overline{b}_i)$ is preintegration measurement that is computed according to the biases $\overline{b}_i$ and $\hat{\Upsilon}_{ij}(\hat{b}_i)$ is preintegration measurement that is computed according to the new estimated biases $\hat{b}_i$. Given the biases update $\hat{b}_i\leftarrow \overline{b}_i+\delta b_i$, the first-order Taylor update of the preintegration measurement under assumption of piecewise constant acceleration is defined as\begin{equation}\label{bias_update_covariance}
\hat{\Upsilon}_{ij}(\hat{ b}_i)=\overline{\Upsilon}_{ij}(\overline{ b}_i)Exp\left(\frac{\partial \hat{\Upsilon}_{ij}}{\partial {\overline{ b}_i}}|_{\overline{ b}_i} \delta  b_i\right)
\end{equation}

If the iterative calculation method of the Jacobian matrix of the preintegration measurement with respect to the bias is considered, we compute
\begin{equation}\label{bias_update_covariance1}
\begin{aligned}
&\hat{\Upsilon}_{i,j+1}(\hat{ b}_i)\overset{(\ref{iterative_update_upsilon_ij})}{=}\Phi_{j,j+1}(\hat{\Upsilon}_{ij}(\hat{ b}_i))\hat{\Upsilon}_{j,j+1}(\hat{ b}_i)\\
=&\Phi_{j,j+1}\left(\overline{\Upsilon}_{ij}(\overline{ b}_i)Exp\left(\frac{\partial \hat{\Upsilon}_{ij}}{\partial {\overline{ b}_i}}|_{\overline{ b}_i} \delta  b_i\right)\right)
\overline{\Upsilon}_{j,j+1}(\overline{ b}_j)Exp\left(\frac{\partial \hat{\Upsilon}_{j,j+1}}{\partial {\overline{ b}_j}}|_{\overline{ b}_j} \delta  b_j\right)\\
\overset{(\ref{automorphism_Phi})}{=}&\Phi_{j,j+1}\left(\overline{\Upsilon}_{ij}(\overline{ b}_i)\right)Exp\left( F\frac{\partial \hat{\Upsilon}_{ij}}{\partial {\overline{ b}_i}}|_{\overline{ b}_i} \delta  b_i\right)\overline{\Upsilon}_{j,j+1}(\overline{ b}_j)Exp\left( G_j \delta b_j\right)\\
=&\Phi_{j,j+1}\left(\overline{\Upsilon}_{ij}(\overline{ b}_i)\right)\overline{\Upsilon}_{j,j+1}(\overline{ b}_j)
\overline{\Upsilon}_{j,j+1}^{-1}(\overline{ b}_j)
Exp\left( F\frac{\partial \hat{\Upsilon}_{ij}}{\partial {\overline{ b}_i}}|_{\overline{ b}_i} \delta b_i\right)\overline{\Upsilon}_{j,j+1}(\overline{  b}_j) Exp\left( G_j \delta b_i\right)\\
=&\Phi_{j,j+1}\left(\overline{\Upsilon}_{ij}(\overline{ b}_i)\right)\overline{\Upsilon}_{j,j+1}(\overline{ b}_j)
Exp\left(Ad_{\overline{\Upsilon}_{j,j+1}^{-1}(\overline{ b}_j)}  F\frac{\partial \hat{\Upsilon}_{ij}}{\partial {\overline{ b}_i}}|_{\overline{ b}_i} \delta  b_i\right) Exp\left( G_j \delta  b_i\right)\\
\overset{(\ref{A_i})}{=}&\Phi_{j,j+1}\left(\overline{\Upsilon}_{ij}(\overline{ b}_i)\right)\overline{\Upsilon}_{j,j+1}(\overline{ b}_j)
Exp\left(  A_j\frac{\partial \hat{\Upsilon}_{ij}}{\partial {\overline{ b}_i}}|_{\overline{ b}_i} \delta  b_i\right) Exp\left( G_j \delta  b_i\right)\\
=&\overline{\Upsilon}_{i,j+1}(\overline{ b}_i)Exp\left(\frac{\partial \hat{\Upsilon}_{i,j+1}}{\partial {\overline{ b}_i}}|_{\overline{ b}_i} \delta  b_i\right)
\end{aligned}
\end{equation}
where the biases are assumed to be constant during the time interval, that is, $\delta b_j=\delta b_i$. 

The recursive calculation formula can be obtained by the first-order approximation of the exponential mapping, that is
\begin{equation}\label{bias_update_recursive_1}
\frac{\partial \hat{\Upsilon}_{i,j+1}}{\partial {\overline{ b}_i}}|_{\overline{ b}_i} =Ad_{\overline{\Upsilon}_{j,j+1}^{-1}(\overline{ b}_j)}  F\frac{\partial \hat{\Upsilon}_{ij}}{\partial {\overline{ b}_i}}|_{\overline{ b}_i} + G_j= A_j \frac{\partial \hat{\Upsilon}_{ij}}{\partial {\overline{ b}_i}}|_{\overline{ b}_i} + G_j
\end{equation}
where the initial value is set as $\frac{\partial \hat{\Upsilon}_{ii}}{\partial {\overline{ b}_i}}|_{\overline{ b}_i}= 0_{9\times 6}$.

On the other hand, we can obtain the Jacobian matrix with respect to the bias at any first, then formulate it as a recursive calculation. We first assume that the perturbation on the bias at time $t_i$ and $t_j$ is as following
\begin{equation}\label{update_biases}
\begin{aligned}
&\hat{\Upsilon}_{ij}(\hat{b}_i)=\begin{bmatrix}
\hat{F}(i,j)(\hat{b}_i^w) & \hat{\mu}_2(i,j)(\hat{b}_i^w,\hat{b}_i^a) & \hat{\zeta}_2(i,j)(\hat{b}_i^w,\hat{b}_i^a)\\
0_{1\times 3} & 1 &0\\
0_{1\times 3} & 0 &1
\end{bmatrix}
=\overline{\Upsilon}_{ij}(\overline{b}_i)Exp\left(\frac{\partial \hat{\Upsilon}_{ij}}{\partial {\overline{b}_i}}|_{\overline{b}_i} \delta b_i\right)\\
=&\begin{bmatrix}
\overline{F}(i,j)(\overline{b}_i^w) & \overline{\mu}_2(i,j)(\overline{b}_i^w,\overline{b}_i^a) & \overline{\zeta}_2(i,j)(\overline{b}_i^w,\overline{b}_i^a)\\
0_{1\times 3} & 1 &0\\
0_{1\times 3} & 0 &1
\end{bmatrix}Exp\left(\frac{\partial \hat{\Upsilon}_{ij}}{\partial {\overline{b}_i}}|_{\overline{b}_i} \delta b_i\right)\\
\triangleq&\begin{bmatrix}
\overline{F}(i,j)(\overline{b}_i^w) & \overline{\mu}_2(i,j)(\overline{b}_i^w,\overline{b}_i^a) & \overline{\zeta}_2(i,j)(\overline{b}_i^w,\overline{b}_i^a)\\
0_{1\times 3} & 1 &0\\
0_{1\times 3} & 0 &1
\end{bmatrix}Exp\left( \begin{bmatrix}
A_{i,j} \\ B_{i,j} \\ C_{i,j}
\end{bmatrix}\right)\\
=&\begin{bmatrix}
\overline{F}(i,j)(\overline{b}_i^w) & \overline{\mu}_2(i,j)(\overline{b}_i^w,\overline{b}_i^a) & \overline{\zeta}_2(i,j)(\overline{b}_i^w,\overline{b}_i^a)\\
0_{1\times 3} & 1 &0\\
0_{1\times 3} & 0 &1
\end{bmatrix}\begin{bmatrix}
Exp(A_{i,j}) & \Gamma_{1}(A_{i,j})B_{i,j} & \Gamma_{1}(A_{i,j})C_{i,j}\\
0_{1\times 3} & 1 &0\\
0_{1\times 3} & 0 &1
\end{bmatrix}\\
\approx &\begin{bmatrix}
\overline{F}(i,j)(\overline{b}_i^w) & \overline{\mu}_2(i,j)(\overline{b}_i^w,\overline{b}_i^a) & \overline{\zeta}_2(i,j)(\overline{b}_i^w,\overline{b}_i^a)\\
0_{1\times 3} & 1 &0\\
0_{1\times 3} & 0 &1
\end{bmatrix}\begin{bmatrix}
Exp(A_{i,j}) & B_{i,j} & C_{i,j}\\
0_{1\times 3} & 1 &0\\
0_{1\times 3} & 0 &1
\end{bmatrix}\\
\end{aligned}
\end{equation}
where $A_{ij}$, $B_{ij}$ and $C_{ij}$ all are small quantities and $\Gamma_{1}(A_{i,j})$ is approximated as $I_{3\times 3}$ in the following derivations.

The derivation of the Jacobian matrix is similar to the one we calculate the propagation of noise of the preintegration measurement. Firstly, we compute 
\begin{equation}\label{F_bias}
\begin{aligned}
&\hat{F}(i,j)(\hat{b}_i^w)=\prod_{k=i}^{j-1}Exp((\hat{\omega}_{ib}^b(k)-\hat{b}_i^w)\Delta t)=\prod_{k=i}^{j-1}Exp((\hat{\omega}_{ib}^b(k)-\overline{b}_i^w-\delta b_i^w)\Delta t)\\
\overset{(\ref{BCH_1})}{=}&\prod_{k=i}^{j-1}Exp((\hat{\omega}_{ib}^b(k)-\overline{b}_i^w)\Delta t)Exp(\Gamma_{1}(-(\hat{\omega}_{ib}^b(k)-\overline{b}_i^w)\Delta t)(-\delta b_i^w\Delta t))\\
=&\prod_{k=i}^{j-1}Exp((\hat{\omega}_{ib}^b(k)-\overline{b}_i^w)\Delta t)\prod_{k=i}^{j-1}Exp\left(-\hat{F}_{k+1,j}^{-1}  \Gamma_{1}(-(\hat{\omega}_{ib}^b(k)-\overline{b}_i^w)\Delta t) \delta b_i^w\Delta t \right)\\
=&\overline{F}(i,j)(\overline{b}_i^w)Exp\left(-\sum_{k=i}^{j-1}  \hat{F}_{k+1,j}^{-1}  \Gamma_{1}(-(\hat{\omega}_{ib}^b(k)-\overline{b}_i^w)\Delta t)\Delta t \delta b_i^w\right)\\
=&\overline{F}(i,j)(\overline{b}_i^w)Exp(A_{i,j})
\end{aligned}
\end{equation}
\begin{equation}\label{mu_i_j_bias}
\begin{aligned}
&\hat{\mu}_2(i,j)(\hat{b}_i^w,\hat{b}_i^a)=\sum_{k=i}^{j-1}\hat{F}(i,k){\Gamma}_1((\hat{\omega}_{ib}^b(k)-\hat{b}_i^w)\Delta t)(\hat{f}_{ib}^b(k)-\hat{b}_i^a)\Delta t\\
\overset{(\ref{BCH_1})}{=}&\sum_{k=i}^{j-1}\overline{F}(i,k)(\overline{b}_i^w)Exp(A_{i,k})
{\Gamma}_1((\hat{\omega}_{ib}^b(k)-\overline{b}_i^w)\Delta t)Exp\left(\Gamma_2(-(\hat{\omega}_{ib}^b(k)-\overline{b}_i^w)\Delta t)(-\delta b_i^w \Delta t )    \right)\\
&(\hat{f}_{ib}^b(k)-\overline{b}_i^a-\delta b_i^a)\Delta t\\
\approx &\sum_{k=i}^{j-1} \left( \overline{F}(i,k)(\overline{b}_i^w){\Gamma}_1((\hat{\omega}_{ib}^b(k)-\overline{b}_i^w)\Delta t)+\overline{F}(i,k)(\overline{b}_i^w)(A_{i,k}\times){\Gamma}_1((\hat{\omega}_{ib}^b(k)-\overline{b}_i^w)\Delta t)\right. \\
&\left.+\overline{F}(i,k)(\overline{b}_i^w){\Gamma}_1((\hat{\omega}_{ib}^b(k)-\overline{b}_i^w)\Delta t)(-\Gamma_2(-(\hat{\omega}_{ib}^b(k)-\overline{b}_i^w)\Delta t)\delta b_i^w \Delta t \times)\right)\\
&(\hat{f}_{ib}^b(k)-\overline{b}_i^a-\delta b_i^a)\Delta t\\
\approx &\sum_{k=i}^{j-1}  \overline{F}(i,k)(\overline{b}_i^w){\Gamma}_1((\hat{\omega}_{ib}^b(k)-\overline{b}_i^w)\Delta t)(\hat{f}_{ib}^b(k)-\overline{b}_i^a-\delta b_i^a)\Delta t\\
&+\overline{F}(i,k)(\overline{b}_i^w)(A_{i,k}\times){\Gamma}_1((\hat{\omega}_{ib}^b(k)-\overline{b}_i^w)\Delta t)(\hat{f}_{ib}^b(k)-\overline{b}_i^a)\Delta t\\
&+\overline{F}(i,k)(\overline{b}_i^w){\Gamma}_1((\hat{\omega}_{ib}^b(k)-\overline{b}_i^w)\Delta t)(-\Gamma_2(-(\hat{\omega}_{ib}^b(k)-\overline{b}_i^w)\Delta t)\delta b_i^w \Delta t )\times(\hat{f}_{ib}^b(k)-\overline{b}_i^a)\Delta t\\
=&\overline{\mu}_2(i,j)(\overline{b}_i^w,\overline{b}_i^a)+\sum_{k=i}^{j-1} -\overline{F}(i,k)(\overline{b}_i^w){\Gamma}_1((\hat{\omega}_{ib}^b(k)-\overline{b}_i^w)\Delta t)\delta b_i^a\Delta t\\
&-\overline{F}(i,k)(\overline{b}_i^w)\left[  {\Gamma}_1((\hat{\omega}_{ib}^b(k)-\overline{b}_i^w)\Delta t)(\hat{f}_{ib}^b(k)-\overline{b}_i^a)\right] \times A_{i,k}\Delta t\\
&+\overline{F}(i,k)(\overline{b}_i^w){\Gamma}_1((\hat{\omega}_{ib}^b(k)-\overline{b}_i^w)\Delta t)(\hat{f}_{ib}^b(k)-\overline{b}_i^a)\times \Gamma_2(-(\hat{\omega}_{ib}^b(k)-\overline{b}_i^w)\Delta t)\delta b_i^w \Delta t^2\\
=&\overline{\mu}_2(i,j)(\overline{b}_i^w,\overline{b}_i^a)+\overline{F}(i,j)(\overline{b}_i^w)B_{i,j}
\end{aligned}
\end{equation}
\begin{equation}\label{zeta_i_j_bias}
\begin{aligned}
&\hat{\zeta}_2(i,j)(\hat{b}_i^w,\hat{b}_i^a)=\sum_{k=i}^{j-1}\left(\hat{F}(i,k)\Gamma_2((\hat{\omega}_{ib}^b(k)-\hat{b}_i^w)\Delta t)(\hat{f}_{ib}^b(k)-\hat{b}_i^a)\Delta t^2+\hat{\mu}_2(i,k)\Delta t \right)\\
\overset{(\ref{BCH_1})}{=}&\sum_{k=i}^{j-1}\overline{F}(i,k)(\overline{b}_i^w)Exp(A_{i,k})
{\Gamma}_2((\hat{\omega}_{ib}^b(k)-\overline{b}_i^w)\Delta t)Exp\left(\Gamma_3(-(\hat{\omega}_{ib}^b(k)-\overline{b}_i^w)\Delta t)(-\delta b_i^w \Delta t )    \right)\\
&(\hat{f}_{ib}^b(k)-\overline{b}_i^a-\delta b_i^a)\Delta t^2+\left( \overline{\mu}_2(i,k)(\overline{b}_i^w,\overline{b}_i^a)+\overline{F}(i,k)(\overline{b}_i^w)B_{i,k}\right)\Delta t\\
\approx &\sum_{k=i}^{j-1} \left( \overline{F}(i,k)(\overline{b}_i^w){\Gamma}_2((\hat{\omega}_{ib}^b(k)-\overline{b}_i^w)\Delta t)+\overline{F}(i,k)(\overline{b}_i^w)(A_{i,k}\times){\Gamma}_2((\hat{\omega}_{ib}^b(k)-\overline{b}_i^w)\Delta t)\right. \\
&\left.+\overline{F}(i,k)(\overline{b}_i^w){\Gamma}_2((\hat{\omega}_{ib}^b(k)-\overline{b}_i^w)\Delta t)(-\Gamma_3(-(\hat{\omega}_{ib}^b(k)-\overline{b}_i^w)\Delta t)\delta b_i^w \Delta t \times)\right)\\
&(\hat{f}_{ib}^b(k)-\overline{b}_i^a-\delta b_i^a)\Delta t^2+\left( \overline{\mu}_2(i,k)(\overline{b}_i^w,\overline{b}_i^a)+\overline{F}(i,k)(\overline{b}_i^w)B_{i,k}\right)\Delta t\\
\approx &\sum_{k=i}^{j-1}  \overline{F}(i,k)(\overline{b}_i^w){\Gamma}_2((\hat{\omega}_{ib}^b(k)-\overline{b}_i^w)\Delta t)(\hat{f}_{ib}^b(k)-\overline{b}_i^a-\delta b_i^a)\Delta t^2\\
&+\overline{F}(i,k)(\overline{b}_i^w)(A_{i,k}\times){\Gamma}_2((\hat{\omega}_{ib}^b(k)-\overline{b}_i^w)\Delta t)(\hat{f}_{ib}^b(k)-\overline{b}_i^a)\Delta t^2\\
&+\overline{F}(i,k)(\overline{b}_i^w){\Gamma}_2((\hat{\omega}_{ib}^b(k)-\overline{b}_i^w)\Delta t)(-\Gamma_3(-(\hat{\omega}_{ib}^b(k)-\overline{b}_i^w)\Delta t)\delta b_i^w \Delta t )\times(\hat{f}_{ib}^b(k)-\overline{b}_i^a)\Delta t^2\\
&+\left( \overline{\mu}_2(i,k)(\overline{b}_i^w,\overline{b}_i^a)+\overline{F}(i,k)(\overline{b}_i^w)B_{i,k}\right)\Delta t\\
=&\overline{\zeta}_2(i,j)(\overline{b}_i^w,\overline{b}_i^a)+\sum_{k=i}^{j-1} -\overline{F}(i,k)(\overline{b}_i^w){\Gamma}_2((\hat{\omega}_{ib}^b(k)-\overline{b}_i^w)\Delta t)\delta b_i^a\Delta t^2\\
&-\overline{F}(i,k)(\overline{b}_i^w)\left[ {\Gamma}_2((\hat{\omega}_{ib}^b(k)-\overline{b}_i^w)\Delta t)(\hat{f}_{ib}^b(k)-\overline{b}_i^a)\right]\times A_{i,k}\Delta t^2\\
&+\overline{F}(i,k)(\overline{b}_i^w){\Gamma}_2((\hat{\omega}_{ib}^b(k)-\overline{b}_i^w)\Delta t)(\hat{f}_{ib}^b(k)-\overline{b}_i^a)\times \Gamma_3(-(\hat{\omega}_{ib}^b(k)-\overline{b}_i^w)\Delta t)\delta b_i^w \Delta t^3\\
&+\left( \overline{F}(i,k)(\overline{b}_i^w)B_{i,k}\right)\Delta t
\end{aligned}
\end{equation}

Therefore, the quantities $A_{i,j}$, $B_{i,j}$, and $C_{i,j}$ in equation (\ref{update_biases}) can be expressed as
\begin{equation}\label{A_I_J}
A_{i,j}=Log\left(\overline{F}(i,j)(\overline{b}_i^w)^{-1} \hat{F}(i,j)(\hat{b}_i^w)  \right)=-\sum_{k=i}^{j-1}  \hat{F}_{k+1,j}^{-1}  \Gamma_{1}(-(\hat{\omega}_{ib}^b(k)-\overline{b}_i^w)\Delta t)\Delta t \delta b_i^w
\end{equation}
\begin{equation}\label{B_I_J}
\begin{aligned}
&B_{i,j}=\overline{F}(i,j)(\overline{b}_i^w)^{-1}\left(\hat{\mu}_2(i,j)(\hat{b}_i^w,\hat{b}_i^a)- \overline{\mu}_2(i,j)(\overline{b}_i^w,\overline{b}_i^a)  \right)\\
=&\sum_{k=i}^{j-1} -\overline{F}(k,j)(\overline{b}_i^w)^{-1}{\Gamma}_1((\hat{\omega}_{ib}^b(k)-\overline{b}_i^w)\Delta t)\delta b_i^a\Delta t\\
&-\overline{F}(k,j)(\overline{b}_i^w)^{-1}\left[{\Gamma}_1((\hat{\omega}_{ib}^b(k)-\overline{b}_i^w)\Delta t)(\hat{f}_{ib}^b(k)-\overline{b}_i^a) \right]\times A_{i,k}\Delta t\\
&+\overline{F}(k,j)(\overline{b}_i^w)^{-1}{\Gamma}_1((\hat{\omega}_{ib}^b(k)-\overline{b}_i^w)\Delta t)(\hat{f}_{ib}^b(k)-\overline{b}_i^a)\times \Gamma_2(-(\hat{\omega}_{ib}^b(k)-\overline{b}_i^w)\Delta t)\delta b_i^w \Delta t^2
\end{aligned}
\end{equation}
\begin{equation}\label{C_I_J}
\begin{aligned}
&C_{i,j}=\overline{F}(i,j)(\overline{b}_i^w)^{-1}\left(\hat{\zeta}_2(i,j)(\hat{b}_i^w,\hat{b}_i^a)- \overline{\zeta}_2(i,j)(\overline{b}_i^w,\overline{b}_i^a)  \right)\\
=&\sum_{k=i}^{j-1} -\overline{F}(k,j)(\overline{b}_i^w)^{-1}{\Gamma}_2((\hat{\omega}_{ib}^b(k)-\overline{b}_i^w)\Delta t)\delta b_i^a\Delta t^2\\
&-\overline{F}(k,j)(\overline{b}_i^w)^{-1}\left[{\Gamma}_2((\hat{\omega}_{ib}^b(k)-\overline{b}_i^w)\Delta t)(\hat{f}_{ib}^b(k)-\overline{b}_i^a)\right]\times A_{i,k}\Delta t^2\\
&+\overline{F}(k,j)(\overline{b}_i^w)^{-1}{\Gamma}_2((\hat{\omega}_{ib}^b(k)-\overline{b}_i^w)\Delta t)(\hat{f}_{ib}^b(k)-\overline{b}_i^a)\times \Gamma_3(-(\hat{\omega}_{ib}^b(k)-\overline{b}_i^w)\Delta t)\delta b_i^w \Delta t^3\\
&+\left( \overline{F}(k,j)(\overline{b}_i^w)^{-1}B_{i,k}\right)\Delta t
\end{aligned}
\end{equation}

Taking into account matrix multiplication, the Jacobian matrix with respect to the biases can be expressed in closed form as
\begin{equation}\label{first_order_biases}
\frac{\partial \hat{\Upsilon}_{ij}}{\partial {\overline{b}_i}}|_{\overline{b}_i}=\begin{bmatrix}
\frac{\partial A_{i,j}}{\partial \delta b_i^w} & 0_{3\times 3}\\
\frac{\partial  B_{i,j}}{\partial \delta b_i^w} & \frac{\partial  B_{i,j}}{\partial \delta b_i^a}\\
\frac{\partial C_{i,j}}{\partial \delta b_i^w} & \frac{\partial  C_{i,j}}{\partial \delta b_i^a}
\end{bmatrix}
\end{equation}
where 
\begin{equation}\label{A_ij_omega}
 \frac{\partial A_{i,j}}{\partial \delta b_i^w}=-\sum_{k=i}^{j-1}  \hat{F}_{k+1,j}^{-1}  \Gamma_{1}(-(\hat{\omega}_{ib}^b(k)-\overline{b}_i^w)\Delta t)\Delta t
\end{equation}
\begin{equation}\label{B_ij_omega}
\begin{aligned}
&\frac{\partial  B_{i,j}}{\partial \delta b_i^w}=\sum_{k=i}^{j-1}-\overline{F}(k,j)(\overline{b}_i^w)^{-1}\left[{\Gamma}_1((\hat{\omega}_{ib}^b(k)-\overline{b}_i^w)\Delta t)(\hat{f}_{ib}^b(k)-\overline{b}_i^a) \right]\times \frac{\partial  A_{i,k}}{\partial \delta b_i^w}\Delta t\\
+&\overline{F}(k,j)(\overline{b}_i^w)^{-1}{\Gamma}_1((\hat{\omega}_{ib}^b(k)-\overline{b}_i^w)\Delta t)(\hat{f}_{ib}^b(k)-\overline{b}_i^a)\times \Gamma_2(-(\hat{\omega}_{ib}^b(k)-\overline{b}_i^w)\Delta t) \Delta t^2
\end{aligned}
\end{equation}
\begin{equation}\label{B_ij_acceleration}
\begin{aligned}
\frac{\partial  B_{i,j}}{\partial \delta b_i^a}=\sum_{k=i}^{j-1} -\overline{F}(k,j)(\overline{b}_i^w)^{-1}{\Gamma}_1((\hat{\omega}_{ib}^b(k)-\overline{b}_i^w)\Delta t)\Delta t
\end{aligned}
\end{equation}
\begin{equation}\label{C_ij_omega}
\begin{aligned}
\frac{\partial  C_{i,j}}{\partial \delta b_i^w}=&\sum_{k=i}^{j-1}-\overline{F}(k,j)(\overline{b}_i^w)^{-1}\left[{\Gamma}_2((\hat{\omega}_{ib}^b(k)-\overline{b}_i^w)\Delta t)(\hat{f}_{ib}^b(k)-\overline{b}_i^a) \right]\times \frac{\partial  A_{i,k}}{\partial \delta b_i^w}\Delta t^2\\
+&\overline{F}(k,j)(\overline{b}_i^w)^{-1}{\Gamma}_2((\hat{\omega}_{ib}^b(k)-\overline{b}_i^w)\Delta t)(\hat{f}_{ib}^b(k)-\overline{b}_i^a)\times \Gamma_3(-(\hat{\omega}_{ib}^b(k)-\overline{b}_i^w)\Delta t) \Delta t^3\\
+&\left( \overline{F}(k,j)(\overline{b}_i^w)^{-1}\frac{\partial  B_{i,k}}{\partial \delta b_i^w}\right)\Delta t
\end{aligned}
\end{equation}
\begin{equation}\label{C_ij_acceleration}
\begin{aligned}
\frac{\partial  C_{i,j}}{\partial \delta b_i^a}=\sum_{k=i}^{j-1} -\overline{F}(k,j)(\overline{b}_i^w)^{-1}{\Gamma}_2((\hat{\omega}_{ib}^b(k)-\overline{b}_i^w)\Delta t)\Delta t^2+\left( \overline{F}(k,j)(\overline{b}_i^w)^{-1}\frac{\partial  B_{i,k}}{\partial \delta b_i^a}\right)\Delta t
\end{aligned}
\end{equation}

It is obvious that the proposed IMU preintegration factor provides a closed-form Jacobian matrix for biases and therefore is more accurate. 
Meanwhile, the calculation of the Jacobian can be written in its recursive form , that is
\begin{equation}\label{first_order_biases_recursive_calculation}
\frac{\partial \hat{\Upsilon}_{ij+1}}{\partial {\overline{ b}_i}}|_{\overline{ b}_i}= A_{j}\frac{\partial \hat{\Upsilon}_{ij}}{\partial {\overline{ b}_i}}|_{\overline{ b}_i}+ G_{j}
\end{equation}
where the result is the same with the result given in equation (\ref{bias_update_recursive_1}).
\section{Preintegrated Measurement Residual and Jacobians}
The residual is given as according to equation (\ref{uncertainty_Lie_group})
\begin{equation}\label{residual} 
\begin{aligned}
&r_{ij}\triangleq\begin{bmatrix}
r_{ij}^C\\ r_{ij}^v \\ r_{ij}^r
\end{bmatrix}=Log\left( {\hat{\Upsilon}_{ij}}^{-1}\Upsilon_{ij} \right)\\
=&Log\left( 
\begin{bmatrix}
({\hat{\Upsilon}}_{ij}^{C})^{-1} \Upsilon_{ij}^C & ({\hat{\Upsilon}_{ij}^C})^{-1} (\Upsilon_{ij}^v-\hat{\Upsilon}_{ij}^v) & ({\hat{\Upsilon}_{ij}^{C}})^{-1} (\Upsilon_{ij}^r-\hat{\Upsilon}_{ij}^r) \\
0_{1\times 3} & 1 &0\\
0_{1\times 3} & 0 &1
\end{bmatrix}
\right)\\
\overset{(\ref{quantities_rotation_velocity_position})}{=}&\left( 
\begin{bmatrix}
Log \left[({\hat{\Upsilon}}_{ij}^{C})^{-1}   (\Gamma_{ij}^CC_i)^{-1}C_j \right]\\
J_{Log \left[({\hat{\Upsilon}}_{ij}^{C})^{-1}   (\Gamma_{ij}^CC_i)^{-1}C_j \right]}^{-1} ({\hat{\Upsilon}_{ij}^C})^{-1} \left(C_i^T\left({\Gamma_{ij}^C}^T(v_j-\Gamma_{ij}^v )-v_i)\right) -\hat{\Upsilon}_{ij}^v \right)\\
  J_{Log \left[({\hat{\Upsilon}}_{ij}^{C})^{-1}   (\Gamma_{ij}^CC_i)^{-1}C_j \right]}^{-1} ({\hat{\Upsilon}_{ij}^{C}})^{-1} \left( C_i^T\left({\Gamma_{ij}^C}^T(r_j-\Gamma_{ij}^r )-(r_i+\Delta t_{ij}v_i)\right) -\hat{\Upsilon}_{ij}^r \right)
\end{bmatrix}
\right)\\
=&\left( 
\begin{bmatrix}
Log \left[({\hat{\Upsilon}}_{ij}^{C})^{-1}   (\Gamma_{ij}^CC_i)^{-1}C_j \right]\\
J_{r_{ij}^C}^{-1} ({\hat{\Upsilon}_{ij}^C})^{-1} \left(C_i^T\left({\Gamma_{ij}^C}^T(v_j-\Gamma_{ij}^v )-v_i)\right) -\hat{\Upsilon}_{ij}^v \right)\\
J_{r_{ij}^C}^{-1} ({\hat{\Upsilon}_{ij}^{C}})^{-1} \left( C_i^T\left({\Gamma_{ij}^C}^T(r_j-\Gamma_{ij}^r )-(r_i+\Delta t_{ij}v_i)\right) -\hat{\Upsilon}_{ij}^r \right)
\end{bmatrix}
\right)
\end{aligned}
\end{equation}
where the preintegration factor has been corrected with respect to biases as~\cite{brossard2021associating} 
\begin{equation}\label{bias_update}
\hat{\Upsilon}_{ij}(\hat{b}_i)=\overline{\Upsilon}_{ij}(\overline{b}_i)Exp\left(\frac{\partial \hat{\Upsilon}_{ij}}{\partial {\overline{b}_i}}|_{\overline{b}_i} \delta b_i\right)
\end{equation}

The Jacobian matrix of the residual with respect to $T_i$ can be calculated by perturbing the residual as follows:
\begin{equation}\label{Jacobian_T_i}
\begin{aligned}
&r_{ij}(T_iExp(\xi_i))=Log\left( {\hat{\Upsilon}_{ij}}^{-1}\left( \Gamma_{ij}\Phi_{ij}(T_iExp(\xi_i))\right)^{-1}T_j \right)\\
=&Log\left( {\hat{\Upsilon}_{ij}}^{-1}\left( \Gamma_{ij}\Phi_{ij}(T_i)Exp(F_{\Delta t_{ij}}\xi_i)\right)^{-1}T_j \right)=Log\left( {\hat{\Upsilon}_{ij}}^{-1}Exp(-F_{\Delta t_{ij}}\xi_i)\left( \Phi_{ij}(T_i)\right)^{-1}\Gamma_{ij}^{-1}T_j \right)\\
=&Log\left( {\hat{\Upsilon}_{ij}}^{-1}\left( \Phi_{ij}(T_i)\right)^{-1}\Gamma_{ij}^{-1}T_j\left(
\left( \Phi_{ij}(T_i)\right)^{-1}\Gamma_{ij}^{-1}T_j\right)^{-1}Exp(-F_{\Delta t_{ij}}\xi_i)\left( \Phi_{ij}(T_i)\right)^{-1}\Gamma_{ij}^{-1}T_j \right)\\
=&Log\left(Exp(r_{ij})Ad_{\left(
\left( \Phi_{ij}(T_i)\right)^{-1}\Gamma_{ij}^{-1}T_j\right)^{-1}}Exp(-F_{\Delta t_{ij}}\xi_i) \right)\\
=&Log\left( Exp(r_{ij})Exp(-Ad_{\left(
	\left( \Phi_{ij}(T_i)\right)^{-1}\Gamma_{ij}^{-1}T_j\right)}^{-1}F_{\Delta t_{ij}}\xi_i) \right)\\
\approx & r_{ij}-J_{-r_{ij}}^{-1}Ad_{\left(
	\left( \Phi_{ij}(T_i)\right)^{-1}\Gamma_{ij}^{-1}T_j\right)}^{-1}F_{\Delta t_{ij}}\xi_i=r_{ij}-J_{-r_{ij}}^{-1}Ad_{\left(
	\Upsilon_{ij}\right)}^{-1}F_{\Delta t_{ij}}\xi_i
\end{aligned}
\end{equation}
where $J_{-r_{ij}}^{-1}$ is defined and approximated as following
\begin{equation}\label{Jacobian_right_definition_approximation}
\begin{aligned}
&J_{-r_{ij}}^{-1}=\begin{bmatrix}
J_{-r_{ij}^C}^{-1} & 0_{3\times 3}& 0_{3\times 3}\\
-J_{-r_{ij}^C}^{-1}Q_{vr}J_{-r_{ij}^C}^{-1}&J_{-r_{ij}^C}^{-1}& 0_{3\times 3}\\
-J_{-r_{ij}^C}^{-1}Q_{pr}J_{-r_{ij}^C}^{-1}& 0_{3\times 3}&J_{-r_{ij}^C}^{-1}
\end{bmatrix}=J_{-r_{ij}^C}^{-1}\begin{bmatrix}
I_{3\times 3} & 0_{3\times 3}& 0_{3\times 3}\\
-Q_{vr}J_{-r_{ij}^C}^{-1}&I_{3\times 3}& 0_{3\times 3}\\
-Q_{pr}J_{-r_{ij}^C}^{-1}& 0_{3\times 3}&I_{3\times 3}
\end{bmatrix}\\
\approx &\begin{bmatrix}
J_{-r_{ij}^C}^{-1} & 0_{3\times 3}& 0_{3\times 3}\\
0_{3\times 3}&J_{-r_{ij}^C}^{-1}& 0_{3\times 3}\\
0_{3\times 3}& 0_{3\times 3}&J_{-r_{ij}^C}^{-1}
\end{bmatrix}=J_{-r_{ij}^C}^{-1}I_{9\times 9}\approx I_{9\times 9}
\end{aligned}
\end{equation}
where the definition and calculation of $Q_{vr}$ and $Q_{pr}$ can be found in~\cite{barfoot2017state,luo2020geometry}.

Therefore, the Jacobian with respect to $T_i$ are computed as
\begin{equation}\label{residual_Jacobian_T_i}
\begin{aligned}
&-J_{-r_{ij}^C}^{-1}\begin{bmatrix}
I_{3\times 3} & 0_{3\times 3}& 0_{3\times 3}\\
-Q_{vr}J_{-r_{ij}^C}^{-1}&I_{3\times 3}& 0_{3\times 3}\\
-Q_{pr}J_{-r_{ij}^C}^{-1}& 0_{3\times 3}&I_{3\times 3}
\end{bmatrix}
\begin{bmatrix}
{\Upsilon_{ij}^{C}}^T & 0_{3\times 3}& 0_{3\times 3}\\
-{\Upsilon_{ij}^{C}}^T\left({\Upsilon_{ij}^{v}}\times \right)&{\Upsilon_{ij}^{C}}^T & 0_{3\times 3}\\
-{\Upsilon_{ij}^{C}}^T\left({\Upsilon_{ij}^{r}}\times \right)& 0_{3\times 3}&{\Upsilon_{ij}^{C}}^T 
\end{bmatrix}\begin{bmatrix}
I_{3\times 3} &  0_{3\times 3}&  0_{3\times 3}\\
 0_{3\times 3}& I_{3\times 3} &  0_{3\times 3}\\
 0_{3\times 3}& \Delta t_{ij}I_{3\times 3}& I_{3\times 3}
\end{bmatrix}\\
%=&-\begin{bmatrix}
%J_{-r_{ij}^C}^{-1}{\Upsilon_{ij}^{C}}^T& 0_{3\times 3}& 0_{3\times 3}\\
%-J_{-r_{ij}^C}^{-1}{\Upsilon_{ij}^{C}}^T\left({\Upsilon_{ij}^{v}}\times \right)&J_{-r_{ij}^C}^{-1}{\Upsilon_{ij}^{C}}^T& 0_{3\times 3}\\
%-J_{-r_{ij}^C}^{-1}{\Upsilon_{ij}^{C}}^T\left({\Upsilon_{ij}^{v}}\times \right)& 0_{3\times 3}&J_{-r_{ij}^C}^{-1}{\Upsilon_{ij}^{C}}^T
%\end{bmatrix}
=&J_{-r_{ij}^C}^{-1}\begin{bmatrix}
-{\Upsilon_{ij}^{C}}^T& 0_{3\times 3}& 0_{3\times 3}\\
Q_{vr}J_{-r_{ij}^C}^{-1}{\Upsilon_{ij}^{C}}^T+{\Upsilon_{ij}^{C}}^T\left({\Upsilon_{ij}^{v}}\times \right)&-{\Upsilon_{ij}^{C}}^T& 0_{3\times 3}\\
Q_{pr}J_{-r_{ij}^C}^{-1}{\Upsilon_{ij}^{C}}^T+{\Upsilon_{ij}^{C}}^T\left({\Upsilon_{ij}^{r}}\times \right)& -\Delta t_{ij} {\Upsilon_{ij}^{C}}^T&-{\Upsilon_{ij}^{C}}^T
\end{bmatrix}\\
\approx & \begin{bmatrix}
-{\Upsilon_{ij}^{C}}^T& 0_{3\times 3}& 0_{3\times 3}\\
{\Upsilon_{ij}^{C}}^T\left({\Upsilon_{ij}^{v}}\times \right)&-{\Upsilon_{ij}^{C}}^T& 0_{3\times 3}\\
{\Upsilon_{ij}^{C}}^T\left({\Upsilon_{ij}^{r}}\times \right)& -\Delta t_{ij} {\Upsilon_{ij}^{C}}^T&-{\Upsilon_{ij}^{C}}^T
\end{bmatrix}
\end{aligned}
\end{equation}

The Jacobian matrix of the residual with respect to $T_j$ can be calculated similarly by perturbing the residual as follows:
\begin{equation}\label{Jacobian_T_j}
\begin{aligned}
&r_{ij}(T_jExp(\xi_j))=Log\left( {\hat{\Upsilon}_{ij}}^{-1}\left( \Gamma_{ij}\Phi_{ij}(T_i)\right)^{-1}T_jExp(\xi_j) \right)\\
=&Log\left(Exp(r_{ij}) Exp(\xi_j) \right)\approx  r_{ij}+J_{-r_{ij}}^{-1}\xi_j
\end{aligned}
\end{equation}

Therefore, the Jacobian with respect to $T_j$ are computed as 
\begin{equation}\label{residual_Jacobian_T_j}
\begin{aligned}J_{-r_{ij}^C}^{-1}\begin{bmatrix}
I & 0_{3\times 3}& 0_{3\times 3}\\
-Q_{vr}J_{-r_{ij}^C}^{-1}&I& 0_{3\times 3}\\
-Q_{pr}J_{-r_{ij}^C}^{-1}& 0_{3\times 3}&I
\end{bmatrix}
\approx &\begin{bmatrix}
J_{-r_{ij}^C}^{-1} & 0_{3\times 3}& 0_{3\times 3}\\
0_{3\times 3}&J_{-r_{ij}^C}^{-1}& 0_{3\times 3}\\
0_{3\times 3}& 0_{3\times 3}&J_{-r_{ij}^C}^{-1}
\end{bmatrix}=J_{-r_{ij}^C}^{-1}I_{9\times 9}\approx I_{9\times 9}
\end{aligned}
\end{equation}

One of the advantages of the proposed preintegration is that the Jacobians derived above are independent of the system's trajectories but only depend on the measurements of IMU. Therefore, the Jacobians reflect better descent directions in the iterative optimization algorithms and result in fewer iterations.

The Jacobian matrix of the residual with respect to $\delta b_i$ can be calculated similarly by perturbing the residual as follows:
\begin{equation}\label{Jacobian_delta_b}
\begin{aligned}
&r\left(\hat{\Upsilon}_{ij}(\hat{b}_i)\right)=Log\left( \left(
\overline{\Upsilon}_{ij}(\overline{b}_i)Exp\left(\frac{\partial \hat{\Upsilon}_{ij}}{\partial {\overline{b}_i}}|_{\overline{b}_i} \delta b_i\right)
\right)^{-1}\left( \Gamma_{ij}\Phi_{ij}(T_i)\right)^{-1}T_j \right)\\
=&Log\left(Exp\left(-\frac{\partial \hat{\Upsilon}_{ij}}{\partial {\overline{b}_i}}|_{\overline{b}_i} \delta b_i\right) \left(
\overline{\Upsilon}_{ij}(\overline{b}_i)
\right)^{-1}\left( \Gamma_{ij}\Phi_{ij}(T_i)\right)^{-1}T_j \right)\\
=&Log\left(Exp\left(-\frac{\partial \hat{\Upsilon}_{ij}}{\partial {\overline{b}_i}}|_{\overline{b}_i} \delta b_i\right) Exp(r_{ij}) \right)=Log\left(Exp(r_{ij}Ad_{r_{ij}}^{-1}Exp\left(-\frac{\partial \hat{\Upsilon}_{ij}}{\partial {\overline{b}_i}}|_{\overline{b}_i} \delta b_i\right) ) \right)\\
\approx & r_{ij}-J_{-r_{ij}}^{-1}Ad_{r_{ij}}^{-1}\frac{\partial \hat{\Upsilon}_{ij}}{\partial {\overline{b}_i}}|_{\overline{b}_i} \delta b_i= r_{ij}-J_{r_{ij}}^{-1}\frac{\partial \hat{\Upsilon}_{ij}}{\partial {\overline{b}_i}}|_{\overline{b}_i} \delta b_i
\end{aligned}
\end{equation}
where $J_{r_{ij}}^{-1}$ is defined and approximated as following
\begin{equation}\label{Jacobian_left_definition_approximation}
J_{r_{ij}}^{-1}=\begin{bmatrix}
J_{r_{ij}^C}^{-1} & 0_{3\times 3}& 0_{3\times 3}\\
-J_{r_{ij}^C}^{-1}Q_{vl}J_{r_{ij}^C}^{-1}&J_{r_{ij}^C}^{-1}& 0_{3\times 3}\\
-J_{r_{ij}^C}^{-1}Q_{pl}J_{r_{ij}^C}^{-1}& 0_{3\times 3}&J_{r_{ij}^C}^{-1}
\end{bmatrix}\approx \begin{bmatrix}
J_{r_{ij}^C}^{-1} & 0_{3\times 3}& 0_{3\times 3}\\
0_{3\times 3}&J_{r_{ij}^C}^{-1}& 0_{3\times 3}\\
0_{3\times 3}& 0_{3\times 3}&J_{r_{ij}^C}^{-1}
\end{bmatrix}=J_{-r_{ij}^C}^{-1}I_{9\times 9}
\end{equation}
where the definition and calculation of $Q_{vl}$ and $Q_{pl}$ can be found in~\cite{barfoot2017state,luo2020geometry}.

Therefore, the Jacobian with respect to $\delta b_i$ are computed as 
\begin{equation}\label{residual_Jacobian_b_i}
\begin{aligned}
&-\begin{bmatrix}
J_{r_{ij}^C}^{-1} & 0_{3\times 3}& 0_{3\times 3}\\
0_{3\times 3}&J_{r_{ij}^C}^{-1}& 0_{3\times 3}\\
0_{3\times 3}& 0_{3\times 3}&J_{r_{ij}^C}^{-1}
\end{bmatrix}\frac{\partial \hat{\Upsilon}_{ij}}{\partial {\overline{b}_i}}|_{\overline{b}_i}\\
\overset{(\ref{first_order_biases})}{=}&-\begin{bmatrix}
J_{r_{ij}^C}^{-1} & 0_{3\times 3}& 0_{3\times 3}\\
0_{3\times 3}&J_{r_{ij}^C}^{-1}& 0_{3\times 3}\\
0_{3\times 3}& 0_{3\times 3}&J_{r_{ij}^C}^{-1}
\end{bmatrix}\begin{bmatrix}
\frac{\partial A_{i,j}}{\partial \delta b_i^w} & 0_{3\times 3}\\
\frac{\partial  B_{i,j}}{\partial \delta b_i^w} & \frac{\partial  B_{i,j}}{\partial \delta b_i^a}\\
\frac{\partial C_{i,j}}{\partial \delta b_i^w} & \frac{\partial  C_{i,j}}{\partial \delta b_i^a}
\end{bmatrix}
=\begin{bmatrix}
-J_{r_{ij}^C}^{-1}\frac{\partial A_{i,j}}{\partial \delta b_i^w} & 0_{3\times 3}\\
-J_{r_{ij}^C}^{-1}\frac{\partial  B_{i,j}}{\partial \delta b_i^w} & -J_{r_{ij}^C}^{-1}\frac{\partial  B_{i,j}}{\partial \delta b_i^a}\\
-J_{r_{ij}^C}^{-1}\frac{\partial C_{i,j}}{\partial \delta b_i^w} & -J_{r_{ij}^C}^{-1}\frac{\partial  C_{i,j}}{\partial \delta b_i^a}
\end{bmatrix}\approx \begin{bmatrix}
\frac{\partial A_{i,j}}{\partial \delta b_i^w} & 0_{3\times 3}\\
\frac{\partial  B_{i,j}}{\partial \delta b_i^w} & \frac{\partial  B_{i,j}}{\partial \delta b_i^a}\\
\frac{\partial C_{i,j}}{\partial \delta b_i^w} & \frac{\partial  C_{i,j}}{\partial \delta b_i^a}
\end{bmatrix}
\end{aligned}
\end{equation}

These analytical expressions for Jacobian matrices of the residual errors are important for the factor graph based optimization.
\section{Monotonicity of Uncertainty Propagation on Matrix Lie Group $SE_2(3)$}
The uncertainty propagation has been extensively studied on matrix Lie group $SE(2)$~\cite{carrillo2015monotonicity, kim2017uncertainty} and $SO(3)$~\cite{rodriguez2018importance}. Kim et al. argued that keeping monotonicity is a matter of correctly modeling errors and propagating uncertainty and proofed that the monotonicity of uncertainty is preserved for A-opt (proportional to the trace of the uncertainty matrix), D-opt (proportional to the determinant of the uncertainty matrix), and E-opt (largest eigenvalue of the uncertainty matrix) using absolute representation of uncertainty~\cite{kim2017uncertainty}. 
The monotonicity is preserved for D-opt criteria and Shannon entropy criteria under a first-order linearized error framework when the absolute representation of uncertainty is used, and is preserved for A-opt, D-opt, E-opt, and Shannon entropy when the differential representation of uncertainty is used~\cite{rodriguez2018importance}.

In this chapter, we will show that the monotonicity is preserved for D-opt and R\'enyi entropy under second order accuracy.

The uncertainty propagation on matrix Lie group $SE_2(3)$ is 
\begin{equation}\label{covariance_recursive_1}
\Sigma_j=A_{i}^{j-1}\Sigma_i {A_{i}^{j-1}}^T+\sum_{k=i}^{j-1}A_{k+1}^{j-1}G_kQG_k^T {A_{k+1}^{j-1}}^T
\end{equation}
The covariance consists two terms where the first term is related to the initial extended pose uncertainty and the second term is related to the motion's uncertainty.
Note that the matrix $A_i$ is an lower triangular matrix with its determinant satisfies $\det A_i=1$. This property means the model can be referred to as a volume preserving process. The matrix $\Phi$ can be seen as matrix representation of the special linear group.

Taking the D-opt in both sides of equation (\ref{covariance_recursive_1}), leads to
\begin{equation}\label{covariance_recursive_1_2}
\det (\Sigma_j)^{\frac{1}{m}}=det\left(A_{i}^{j-1}\Sigma_i {A_{i}^{j-1}}^T+\sum_{k=i}^{j-1}A_{k+1}^{j-1}G_kQG_k^T {A_{k+1}^{j-1}}^T \right)^{\frac{1}{m}}
\end{equation}
Using the Minkowski's inequality, it follows that
\begin{equation}\label{covariance_recursive_1_3}
\det (\Sigma_j)^{\frac{1}{m}}\geq det\left(A_{i}^{j-1}\Sigma_i {A_{i}^{j-1}}^T \right)^{\frac{1}{m}}
+det\left(\sum_{k=i}^{j-1}A_{k+1}^{j-1}G_kQG_k^T {A_{k+1}^{j-1}}^T \right)^{\frac{1}{m}}
\end{equation}
Since $\det A_i=1$, we can obtain that $\det A_i^{j-1}=1$. Therefore, we have 
\begin{equation}\label{covariance_recursive_1_4}
\det (\Sigma_j)^{\frac{1}{m}}\geq det\left(\Sigma_i  \right)^{\frac{1}{m}}
+det\left(\sum_{k=i}^{j-1}G_kQG_k^T  \right)^{\frac{1}{m}}
\end{equation}

Meanwhile, as the matrix $Q$ is symmetric positive semidefinite, its determinant is non-negative and the uncertainty $G_kQG_k^T$ is also semidefinite positive definite because
\begin{equation}\label{semedefinite}
x^TG_kQG_k^Tx=(G_k^Tx)^TQ(G_k^Tx)\geq 0
\end{equation}
holds for all $x\in \mathbb{R}^9$.

Finally, we can obtain that 
\begin{equation}\label{covariance_recursive_1_5}
\det (\Sigma_j)^{\frac{1}{m}}\geq det\left(\Sigma_i  \right)^{\frac{1}{m}}
+det\left(\sum_{k=i}^{j-1}G_kQG_k^T  \right)^{\frac{1}{m}}\geq det\left(\Sigma_i  \right)^{\frac{1}{m}}
\end{equation}
Therefore, 
\begin{equation}\label{}
\det(\Sigma_j)\geq \det(\Sigma_i)
\end{equation} 

The R\'enyi entropy of the multivariate Gaussian distribution is given as
\begin{equation}\label{renyi_entropy_mg}
H_{\alpha}=\frac{1}{2}\log\det \left(2\pi\alpha^{\frac{1}{\alpha-1}} P \right)=\frac{N}{2}\log (2\pi\alpha ^{\frac{1}{\alpha-1}})+\frac{1}{2}\log(\det P)
\end{equation}
where $\alpha\in[0,1)\cup (1,\infty)$ is a free parameter providing a family of entropy functionals, $N$ is the dimension of the state and $P$ is the associated covariance matrix which represents the uncertainty. The Shannon entropy of the multivariate Gaussian distribution can be obtained as $\alpha\rightarrow 1$.

As the monotonicity of the Shannon entropy is equivalent to that of the D-opt optimality criteria, the monotoicity of the R\'enyi entropy is also equivalent to the monotonicity of the D-opt optimality criteria because the uncertainty of R\'enyi entropy is differs from the uncertainty of Shannon entropy only on the free parameter $\alpha$. This can be shown as follows
\begin{equation}\label{renyi_entropy_uncertainty}
\begin{aligned}
H_{\alpha}(\Sigma_j)-H_{\alpha}(\Sigma_i)=\frac{1}{2}\log \left(\frac{\det \Sigma_j}{\det \Sigma_i}\right)\geq 0
\end{aligned}
\end{equation}
%In order to improve the efficacy of the learning based inertial model, the raw IMU data are replaced by the preintegrated features~\cite{khorrambakht2020preintegrated, khorrambakht2021deep}.
%%%%%%%%%%%%%%%%%%%%%
\section{Conclusions}
In this paper, a unified mathematical framework for IMU preintegration in inertial-integrated system is proposed. This framework aims to exactly discretize the system state as a global navigation state, a global increment, and a local increment.
The calculation for global increment in transformed navigation frame is first proposed and derived.
This procedure can be applied to the global increment in navigation frame and ECEF frame too.
As concerns the local increment, the algorithms based on local acceleration and global acceleration are summarized. Most important, however, is that we consider the equivariant rotation vector update algorithms in traditional mechanization can be adopted to the calculation of local increment according to the motion of the agent, which makes the IMU preintegration methods can be applied to various grade IMU in various frames under different motion environment.
In the future, more experiments should be performed to show the performance of the proposed mathematical framework under different situation.

\vspace{2ex}
%%%%%%%%%%%%%%%%%%%%%%%%%%%%%%%%%%%%%%%%%%%%%%%%%%%%%%%%%%%%%%%%%%%%%%%%%%%%%%%%%%%%%%%%%%%%%%%%%%%%%%%%%%%%%%%%%%%%%
%%%%%%%%%%%%%%%%%%%%%%%%%%%%%%%%%%%%%%%%%%%%%%%%%%%%%%%%%%%%%%%%%%%%%%%%%%%%%%%%%%%%%%%%%%%%%%%%%%%%%%%%%%%%%%%%%%%%%
\noindent
{\bf\normalsize Acknowledgement}\newline
{This research was supported by a grant from the National Key Research and Development Program of China (2018YFB1305001).} 
%We express thanks to professor Xiaoji Niu from the GNSS Research Center, Wuhan University.} \vspace{2ex}

\bibliographystyle{IEEEtran}
%Bibliography file should be compiled with BibTex
\bibliography{ref.bib}

\end{document}